\documentclass{article} % For LaTeX2e
\usepackage{iclr2027_conference,times}
\usepackage{graphicx} % Required for inserting images

\usepackage{amsmath,amsfonts,bm}

\def\eqref#1{equation~\ref{#1}}
\def\1{\bm{1}}

\DeclareMathAlphabet{\mathsfit}{\encodingdefault}{\sfdefault}{m}{sl}
\SetMathAlphabet{\mathsfit}{bold}{\encodingdefault}{\sfdefault}{bx}{n}

\usepackage{hyperref}
\usepackage{url}
\usepackage{algorithm}
\usepackage[noend]{algpseudocode}
\usepackage{booktabs}
\usepackage{xcolor}
\usepackage{wrapfig}
\usepackage{placeins}
\usepackage{listings}
\usepackage{flafter} % Keep floats after their source position.
\usepackage{enumitem}
\hypersetup{hypertexnames=false}

\title{
Mixture of Self-Improving Branches\\
for Agent Harness Optimization}

\author{%
\textbf{Haoyu Dong\textsuperscript{1,2,*}\quad Yuhang Zhou\textsuperscript{1}\quad Zihao Lin\textsuperscript{1,3,*}\quad Yifan Wu\textsuperscript{1}\quad Bo Peng\textsuperscript{1}}\\[3pt]
\textbf{Mingyi Wang\textsuperscript{1}\quad Xiangjun Fan\textsuperscript{1}\quad Lizhu Zhang\textsuperscript{1,\dag}\quad Zhuokai Zhao\textsuperscript{1,\dag}}}

\iclrfinalcopy % Uncomment for camera-ready version, but NOT for submission.
\begin{document}

\maketitle
\begingroup
\renewcommand{\thefootnote}{}
\makeatletter
\long\def\@makefntext#1{\noindent#1}
\makeatother
\footnotetext{\textsuperscript{1}Meta \quad \textsuperscript{2}Duke University \quad \textsuperscript{3}University of California, Davis.\\
\textsuperscript{*}Work done while at Meta.
\textsuperscript{\dag}Co-last authors.
Correspondence to: Haoyu Dong \textless\href{mailto:haoyu.dong151@duke.edu}{haoyu.dong151@duke.edu}\textgreater\ and Zhuokai Zhao \textless\href{mailto:zhuokai@meta.com}{zhuokai@meta.com}\textgreater.}
\endgroup

\begin{abstract}
%An LLM agent's capability depends not only on the underlying model but also on the surrounding \textit{harness}, which coordinates its reasoning, tool use, and execution.
Harness optimization provides a practical setting for recursive self-improvement (RSI), where agent-generated modifications inform subsequent changes through execution feedback.
Recent work such as Meta-Harness implements this process through iterative code generation and evaluation, but retains a fixed development set and proposal policy. 
These constraints channel evolution along a single search trajectory, increasing the risk of converging to a local optimum.
We make the improvement process itself adaptive by organizing search into branches with evolving development subsets and proposal policies.
Each branch retains development cases solved by more of its leading harnesses than by those of other branches, drops cases solved by every leading harness across all branches, and revises its proposal policy using its own search history.
To deploy the resulting complementary harnesses, we propose a router to select one development-selected branch head for each new input before execution.
Across mathematical reasoning and agentic coding benchmarks, our system achieves relative improvements over Meta-Harness of 34.8\% on Olympiad-level mathematical reasoning, 11.6\% on Terminal-Bench 2.0, and 3.8\% on SWE-bench Lite, with harness selection and router configuration based solely on development data.
These results show that evolving branch objectives and proposal policies can yield complementary harnesses whose strengths a router combines without access to test outcomes.

%Harness optimization provides a practical setting for recursive self-improvement (RSI), where agent-generated modifications inform subsequent changes through execution feedback.
%Recent work such as Meta-Harness implements this process through iterative code generation and evaluation, but retains a fixed development objective and proposal policy.
%These constraints can repeatedly favor similar harness designs, leaving complementary strengths underexplored.
%We make the improvement process itself adaptive by organizing search into branches with evolving development objectives and proposal policies.
%Each branch retains development cases where its leading candidates have a comparative advantage and removes cases consistently solved across branches.
%Its proposal guidance evolves from local search experience, allowing branches to develop distinct exploration priorities.
%We use a router to select among locally development-best branch harnesses for each new problem before execution.
%Across mathematical reasoning and agentic coding benchmarks, our system achieves relative improvements over Meta-Harness of 34.8\% on Olympiad-level math questions, 11.6\% on Terminal-Bench 2.0, and 3.8\% on SWE-bench Lite, with harness selection and router configuration based solely on development data.
%These results show that adaptive search objectives and proposal policies can discover complementary harnesses whose strengths support improved deployment performance.
\end{abstract}

\section{Introduction}

An LLM agent's capability depends not only on the underlying model but also on its surrounding \textit{harness}.
Through retrieval, tool interfaces, and control flow, the harness governs what information the model receives, which actions it can take, and when it verifies or revises its work \citep{lewis2020rag,yao2023react,shinn2023reflexion}.
Prior work demonstrates that carefully designed reasoning and feedback mechanisms can improve agent performance on question answering, sequential decision-making, and coding tasks \citep{yao2023react,shinn2023reflexion}.
Building on these advances, recent work automates harness design by using evaluation feedback to refine prompts, code, and workflows \citep{khattab2024dspy,hu2024adas}.
For example, Meta-Harness uses an agentic proposer to generate harness implementations and evaluate them on a predefined development set, drawing on prior code, scores, and execution traces to guide subsequent proposals \citep{lee2026metaharness}.
This process provides a practical setting for recursive self-improvement (RSI), where experience from earlier implementations guides subsequent changes.

Although this feedback guides successive harness proposals, the development set and proposal instructions remain fixed throughout search \citep{hu2024adas,lee2026metaharness}.
Evaluating candidates on the same development cases throughout search can overlook complementary strengths: a harness may solve cases missed by the leading ones yet rank lower overall, leaving those capabilities with limited opportunity for further development.
Meanwhile, fixed proposal instructions do not explicitly adapt exploration priorities to the strengths and unresolved failures that emerge during search.
These limitations motivate our central question: can adapting each branch’s development subset and proposal guidance help discover complementary harnesses and improve deployment performance?

\begin{figure}[t]
\centering
\includegraphics[width=\textwidth]{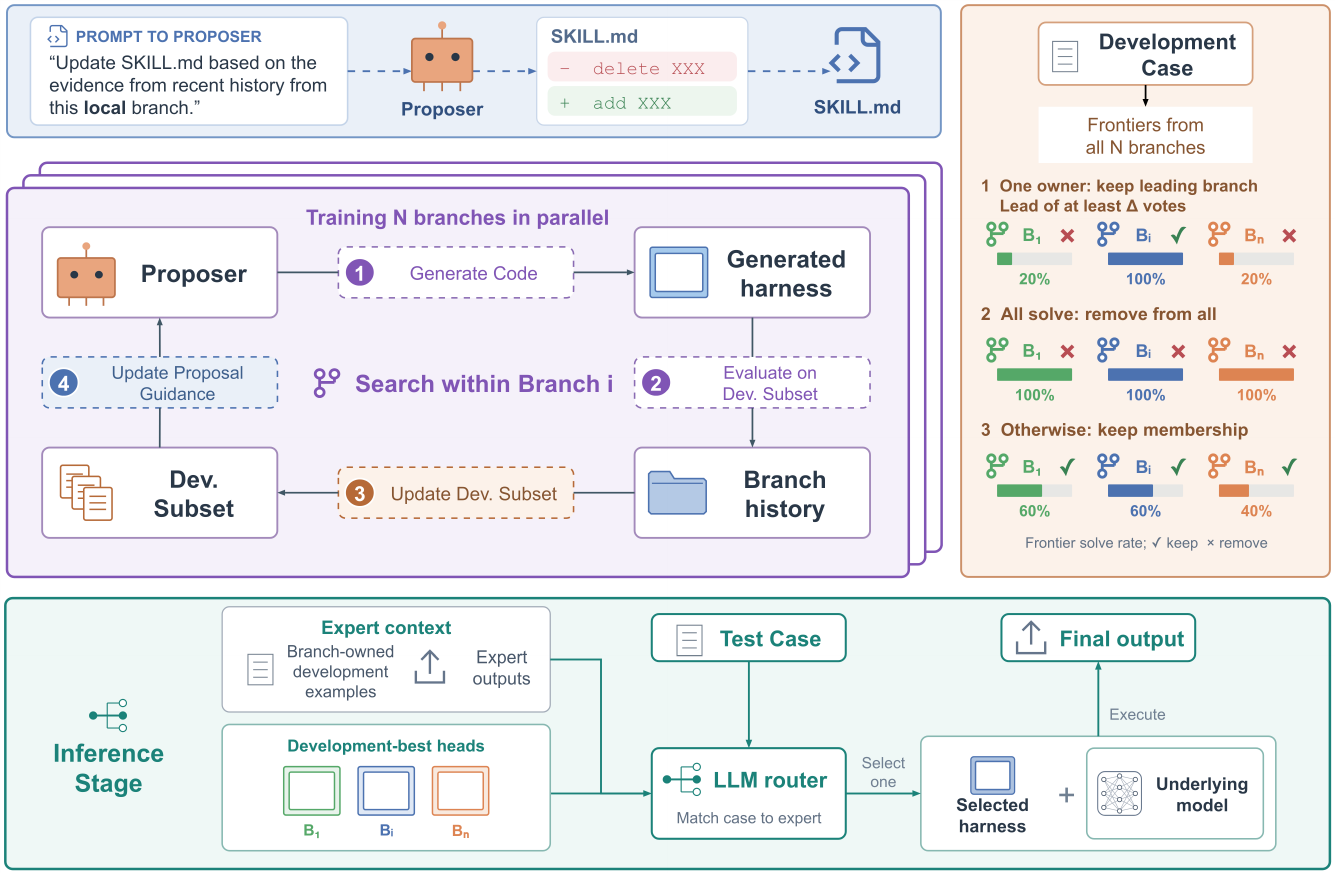}
\vspace{-0.25in}
\caption{\textbf{Pipeline overview.}
Left: each branch proposes and evaluates harnesses using its current development subset, branch history, and proposal guidance.
Periodic guidance updates (upper inset) translate local search experience into priorities for subsequent proposals.
Right: development subsets are updated by comparing how many leading harnesses in each branch solve each case.
Bottom: a router selects one development-best branch harness for each new input before execution.}
\label{fig:branch-meta-harness-overview}
\vspace{-0.15in}
\end{figure}

To answer this question, we propose organizing harness search into separate branches, each with a development subset that adapts to its emerging strengths (Figure~\ref{fig:branch-meta-harness-overview}).
A branch is a separate search process with its own development subset, candidate history, and proposal guidance.
For each development case, we compare how many leading harnesses in each branch solve it.
When one branch has a clear advantage over the others, we retain the case in that branch's subset and remove it from the others.
Cases solved by every leading harness across all branches are removed from search.
By changing which cases contribute to each branch's score, these updates can promote previously overlooked harnesses and redirect subsequent proposals.
To adapt proposal generation to the cases retained by each branch, we also revise each branch's guidance using its local search experience \citep{yang2026skillopt}.
The guidance records which modifications improved performance, which repeatedly failed, and which problems remain unresolved.
These lessons guide subsequent proposals by identifying promising mechanisms, recurring failure modes, and directions for further exploration.
As branches accumulate different evidence, their guidance develops distinct priorities that complement the specialization induced by their development subsets.

Branch-local search makes selecting a single harness for deployment nontrivial.
Selecting by test accuracy requires held-out outcomes that are unavailable when deploying on new problems, while development scores are not directly comparable across branches with different subsets.
To address this challenge, we retain the development-best harness from each branch and use a router to choose among them for each new problem, inspired by mixture-of-experts and model-routing approaches \citep{jacobs1991mixtures,ong2024routellm}.

We evaluate our approach on Olympiad-level mathematical reasoning \citep{lee2026metaharness}, Terminal-Bench 2.0 \citep{merrill2026terminalbench}, and SWE-bench Lite \citep{jimenez2024swebench}.
With harnesses selected exclusively on development data, the routed system improves over Meta-Harness across all four benchmark--model settings.
On mathematical reasoning, accuracy rises from 46.0\% to 62.0\% with Gemini 3 Flash and from 29.0\% to 30.5\% with Claude Sonnet 4.5, corresponding to relative gains of 34.8\% and 5.2\%.
The corresponding relative gains on Terminal-Bench 2.0 and SWE-bench Lite are 11.6\% and 3.8\%, reaching 50.0\% task completion and 66.0\% issue resolution, respectively.
The router achieves performance comparable to the branch expert with the highest test accuracy and surpasses it in two settings, without using test outcomes for selection.
Trajectory analysis illustrates how changing development subsets redirects search, while ablations show the strongest performance when development updates and proposal adaptation are combined.

Our contributions are fourfold:
\begin{itemize}[topsep=0pt,leftmargin=*,noitemsep]
    \item \textbf{Search with dynamic development subsets.}
Multiple search branches adapt their development subsets based on differences in task performance, encouraging complementary harnesses.
    \item \textbf{Branch-specific proposal adaptation.}
Proposal guidance evolves from each branch's local search history, preserving distinct lessons and priorities for future modifications.
    \item \textbf{Deployable expert routing.}
A router combines complementary development-selected branch experts, approaching or surpassing the test accuracy of the best individual expert.
    \item \textbf{Improved benchmark performance.}
Development-selected systems outperform fixed baselines and Meta-Harness across four settings spanning mathematical reasoning and agentic coding.

\end{itemize}

\section{Related Work}

\noindent\textbf{Self-evolving agents and harnesses.}
Automated agent design searches over executable programs and workflows, making agent implementations an object of optimization \citep{hu2024adas,zhang2024aflow}.
Meta-Harness extends this direction to harness code, using an agentic proposer that inspects prior implementations, scores, and execution traces through a persistent filesystem \citep{lee2026metaharness}.
It provides the search framework and primary baseline for our work.
Self-Harness identifies failure patterns from execution traces, proposes targeted harness modifications, and validates them through regression testing \citep{zhang2026selfharness}.
Other work explores automated harness optimization, online adaptation, and joint evolution of system components \citep{sengupta2026harbor,karten2026continualharness,chen2026coharness,chen2026harnessforge,luo2026autodesign,hao2026self}.
These approaches establish harness evolution as a practical form of agent self-improvement.
Our contribution concerns how that evolution is directed: multiple branches adapt their development subsets through comparative frontier coverage, changing which candidate harnesses are favored and extended.

\noindent\textbf{Diversity through adaptive search inputs.}
LLM-driven evolutionary systems encourage novelty through diverse candidate archives, drawing on novelty and quality-diversity search \citep{lehman2011novelty,mouret2015mapelites}.
AlphaEvolve \citep{novikov2025alphaevolve} preserves alternative programs in a MAP-Elites-inspired database.
AgenticGEO \citep{yuan2026agenticgeo} retains diverse rewriting strategies in behavioral cells.
These mechanisms promote diversity through candidate scoring and retention.
We instead adapt each branch's development set through comparative frontier coverage, changing the cases that guide selection and modification.
This encourages complementary harnesses through different search objectives while keeping the task-level scoring rule fixed.

\noindent\textbf{Evolving guidance from search experience.}
Another direction for improving self-evolution is to use accumulated experience to refine the guidance for future changes.
Prompt and program optimizers revise instructions and programs using execution feedback \citep{pryzant2023protegi,yang2024opro,yuksekgonul2024textgrad,khattab2024dspy}.
DREvo \citep{guo2026drevo} distills historical evidence into harness-search directions, while SkillOpt \citep{yang2026skillopt} updates external skill guidance from scored rollouts.
We use branch-specific guidance to turn local successes and unresolved failures into distinct proposal priorities as objectives diverge.

\noindent\textbf{Deploying evolved harnesses.}
Recent harness-evolution studies often use overlapping tasks for search and final evaluation \citep{lee2026metaharness,wang2026rethinking}.
This practice can overstate generalizable improvements, with reported gains on search tasks largely failing to transfer to held-out tasks \citep{wang2026rethinking}.
We therefore evaluate deployable performance, selecting harnesses on development data alone.
Selection is less straightforward in our setting because scores on different branch-specific development subsets are not directly comparable.
To solve this, we route each input to a development-best branch harness, following input-level model selection \citep{chen2023frugalgpt,ong2024routellm}.
The broader principle of expert selection \citep{jacobs1991mixtures} also appears in model-internal routing \citep{zeng2026s,feng2026dag} and token-level collaboration \citep{xiong2026token}.
Routing approaches or exceeds the stronger branch harness without test-based selection.

\section{Branching Search with Evolving Objectives and Policies}
\label{sec:method}

\noindent\textbf{Overview.}
We organize harness evolution into multiple search branches.
Following~\citet{lee2026metaharness}, we use Claude Opus 4.6 \citep{anthropic2026opus46} as the coding proposer.
During search, we adapt each branch's development subset by comparing task performance across branches and periodically revise its proposal guidance using local search experience.
At deployment, a router selects one branch expert for each new problem.

\noindent\textbf{Branch-local search and proposal generation.}
A branch is a separate search process that evolves harnesses using its own development subset, candidate history, and proposal guidance.
Let $H$ denote a candidate harness, $M$ the action language model, and $P$ the coding proposer shared across branches.
At iteration $t$, let $\mathcal{X}_b^t\subseteq\mathcal{X}$ denote branch $b$'s current subset of the full development set $\mathcal{X}$.
% \yz{Please explicitly distinguish the candidate population $\mathcal{H}_b^t$, which contains harness implementations?}
The candidate population $\{H\}_b^t$ contains the branch's harness implementations, whose source code, evaluation scores, and execution traces are stored in its local filesystem $\mathcal{D}_b^t$.
The branch's proposal guidance $S_b^t$ is stored in \texttt{SKILL.md}.
All branches are initialized with the same seed harnesses $\{H\}^0$, full development set $\mathcal{X}$, and proposal guidance $S^0$.

Executing $H$ with $M$ on instance $x$ produces a potentially stochastic trajectory $f_M(H,x)$ with task reward $r(f_M(H,x),x)$.
Branch $b$ seeks to maximize the expected reward on $\mathcal{X}_b^t$:
\begin{equation}
J_b^t(H)
= \frac{1}{|\mathcal{X}_b^t|}
  \sum_{x\in\mathcal{X}_b^t}
  \mathbb{E}\!\left[r(f_M(H,x),x)\right],
\label{eq:branch-objective}
\end{equation}
where the expectation is over execution randomness.
We estimate this objective using the mean observed reward and select the highest-scoring candidate in $\{H\}_b^t$ as the branch head $\widehat{H}_b^t$.

% \yz{Should we index the proposed harness by both branch and iteration, for example $H_b^{t+1}$, to distinguish it from a generic harness $H$ and the selected head $\widehat{H}_b^t$? Please align the indexing with the algorithm.}
At each iteration, $P$ first reads the harness--score pairs $(H,J_b^t(H))$ for existing candidates in $\{H\}_b^{t-1}$.
Within its own branch's filesystem $\mathcal{D}_b^t$, $P$ is free to explore execution traces and other artifacts, deciding which evidence to inspect and which failures to investigate under guidance $S_b^t$.
Using this evidence, $P$ proposes a new harness $H_b^t$ and records its design rationale.
We evaluate the candidate on $\mathcal{X}_b^t$ and add its implementation and evaluation records to $\mathcal{D}_b^t$ for subsequent proposals.

\begin{algorithm}[t]
\caption{Harness search with evolving objectives and proposal guidance}
\label{alg:trajectory-evolving-meta-harness}
\begin{algorithmic}[1]
\State \textbf{Input:} tasks $\mathcal{X}$, model $M$, proposer $P$, branches $B$, iterations $N$, guidance $S^0$
\State Select and evaluate seed harnesses $\{H\}^0$ on $\mathcal{X}$, storing records in $\mathcal{D}^0$
\For{$b=1,\ldots,B$}
  \State $\{H\}_b^0\leftarrow\{H\}^0$, $\mathcal{D}_b^0\leftarrow\operatorname{Copy}(\mathcal{D}^0)$, $\mathcal{X}_b^0\leftarrow\mathcal{X}$, $S_b^0\leftarrow S^0$
\EndFor
\For{$t=1,\ldots,N$}
  \ForAll{branches $b$ \textbf{in parallel}}
    \State $H_b^t \leftarrow P\!\left(\{(H,J_b^t(H)):H\in\{H\}_b^{t-1}\},\mathcal{D}_b^t,\mathcal{X}_b^t,S_b^t\right)$
    \State Evaluate $H_b^t$ on $\mathcal{X}_b^t$ and store $(H_b^t,J_b^t(H_b^t),\text{traces})$ in $\mathcal{D}_b^t$
    \State $\{H\}_b^t\leftarrow\{H\}_b^{t-1}\cup\{H_b^t\}$
  \EndFor
  \If{$t \bmod C_p = 0$ and $\min_b |\{H\}_b^t| \ge q$}
    \State $\mathcal{F}_b^t\leftarrow\Call{TopQ}{\{H\}_b^t,J_b^t,q}$
    for each branch $b$
    \State $\{\mathcal{X}_b^t\}_{b=1}^B\leftarrow
    \Call{PruneDevelopment}{\{\mathcal{X}_b^t,\mathcal{F}_b^t\}_{b=1}^B,\Delta}$
  \EndIf
  \If{$t\bmod C_s=0$}
    \State $S_b^t\leftarrow\Call{UpdateGuidance}{P,S_b^t,\operatorname{Recent}_{C_s}(\mathcal{D}_b^t)}$
    for each branch $b$
  \EndIf
\EndFor
\State Select $\widehat{H}_b^N\in\arg\max_{H\in\{H\}_b^N}J_b^N(H)$ for each branch $b$
\State Build router $R$ from heads $\widehat{H}_b^N$, labeled development cases, and expert outputs
\State \Return $(R,\{\widehat{H}_b^N\}_{b=1}^B)$
\Comment{route each new input before execution}
\end{algorithmic}
\end{algorithm}

\noindent\textbf{Development-driven trajectory evolution.}
The development subset $\mathcal{X}_b^t$ determines which harnesses are favored within branch $b$, allowing us to redirect search by changing the problems on which candidates are evaluated.
We update this subset by comparing performance on individual cases across branch frontiers $\mathcal{F}_b^t$ to determine which cases each branch retains.
The frontier $\mathcal{F}_b^t$ is defined as the top-$q$ harnesses in branch $b$ ranked by mean reward on $\mathcal{X}_b^t$.
Note that because $\mathcal{X}_b^t$ changes during search, candidate rankings and frontier membership can change even when the harness implementations remain unchanged.

Concretely, for each development case $x$, we first compute the coverage vote $c_b^t(x)$ on each branch's frontier $\mathcal{F}_b^t$, counting how many harnesses solve the case:
\begin{equation}
c_b^t(x)=\sum_{H\in\mathcal{F}_b^t}
\mathbb{1}\!\left[H\text{ solves }x\right].
\label{eq:coverage-vote}
\end{equation}
% \yz{Please keep the harness indexing consistent with the preceding subsection. Here, an unindexed $H$ is appropriate as a summation variable over stored harnesses; $\mathcal{F}_b^t$ already identifies the branch and current iteration. Distinguish this usage from the indexed newly proposed harness.}
% \yz{The rule is clear, but could we consolidate it into three cases: remove universally solved cases; assign cases exclusively to a branch with a unique lead of at least $\Delta$ votes; otherwise preserve the existing subset assignments?}
% \yz{Does this mean cases that one branch's frontier solves more consistently than the others? Please state this concretely and distinguish relative coverage from absolute task difficulty.}
If $c_b^t(x)=|\mathcal{F}_b^t|$ for every branch $b$, i.e., every harness in every branch's frontier solves $x$, we remove $x$ from all subsets.
This removes cases that offer little signal for further improvement.
For the remaining cases, branch $b$ gains ownership of $x$ when $c_b^t(x)-\max_{b'\ne b}c_{b'}^t(x)\ge\Delta$, where $\Delta$ is the ownership margin.
We remove these cases from all other branches' subsets, reserving them for the branch with a relative advantage.
Otherwise, subset memberships remain unchanged.
Retaining cases where a branch has a relative advantage gives subsequent search a distinct focus, allowing each branch to build on its emerging strengths without predefined task categories.

\noindent\textbf{Branch-specific proposal adaptation.}
As development subsets $\mathcal{X}_b^t$ diverge, the shared initial guidance $S^0$ may no longer address the challenges specific to each branch.
Inspired by SkillOpt \citep{yang2026skillopt}, we make $S_b^t$ adaptable, allowing each branch to refine its proposal strategy as its development objective evolves.
% \yz{Please define the skill optimizer: which model implements it, what inputs it receives, and whether it is the same model as proposer $P$. Briefly describe its revision instructions here and provide the full prompt in the appendix for reproducibility.}
Specifically, the same proposer $P$ updates $S_b^t$ from the current guidance and recent search records using the prompts reproduced in Appendix~\ref{app:skill-update-prompt}.
The update instructions direct the proposer to compare candidates within the same development stage and distinguish repeated evidence across implementations from isolated successes or failures.
Development updates change which capabilities are rewarded, while guidance updates translate local successes and failures into priorities for subsequent proposals.
A mechanism that helps solve one branch's retained cases can receive continued attention there, while another branch targets different failures.

\begin{samepage}
\noindent\textbf{Development-informed routing.}
% \yz{Could we shorten this motivation to a direct statement that harness selection and router configuration use only development data? This is standard evaluation practice; the distinction from retrospective test-best selection could go in the experimental setup, leaving this subsection focused on the routing mechanism.}
The resulting branches may develop complementary harnesses, but their scores on different development subsets are not directly comparable.
At the end of search, we select the head $\widehat{H}_b^N$ with the highest mean reward on each branch's final development subset $\mathcal{X}_b^N$, retaining one expert per branch.\par
\end{samepage}

To choose among these experts, we use a router $R$, following mixture-of-experts and model-routing approaches \citep{jacobs1991mixtures,chen2023frugalgpt,ong2024routellm}.
We construct labeled development examples from cases solved by exactly one final branch head, assigning the successful head as the routing label.
$R$ uses the action model $M$ with expert source code, these head-exclusive cases, and both heads' outputs (Appendix~\ref{app:router-instructions}).
We optimize its instruction using GEPA \citep{agrawal2025gepa}, with head-exclusive development cases as training data.
For each new problem, $R$ selects a head before execution without expert outputs or correctness feedback.

\noindent\textbf{Full search and deployment pipeline.}
Our full pipeline is described in Algorithm~\ref{alg:trajectory-evolving-meta-harness}.
Branches evolve in parallel at each iteration.
Pruning begins only after every branch has at least $q$ candidate harnesses, after which we compare frontiers and update the development subsets every $C_p$ iterations.
Every $C_s$ iterations, $P$ also revises $S_b^t$.
When both updates are due, we update the development subsets before revising proposal guidance and use both updated states in the next iteration.
After $N$ iterations, we construct router $R$ from the final branch heads and their development evidence.

% Keep the main results tables on separate pages.
\setcounter{topnumber}{1}
\setcounter{totalnumber}{1}

\begin{table}[!t]
\centering
\caption{Held-out performance (\%) of fixed baselines and development-selected systems.
Bold indicates the best score in each setting.}
\label{tab:main-results}
\small
\setlength{\tabcolsep}{6pt}
\renewcommand{\arraystretch}{1.12}
\begin{tabular*}{\textwidth}{@{\extracolsep{\fill}}ccccc@{}}
\toprule
\textbf{Method} & \multicolumn{2}{c}{\textbf{Math}} & \textbf{Terminal-Bench 2.0} & \textbf{SWE-bench Lite} \\
\cmidrule(lr){2-3}\cmidrule(lr){4-4}\cmidrule(l){5-5}
& \shortstack{Gemini 3\\Flash} & \shortstack{Claude Sonnet\\4.5} & \shortstack{Claude Sonnet\\4.5} & \shortstack{Claude Sonnet\\4.5} \\
\midrule
No-memory & 34.0 & 24.5 & --- & --- \\
Few-shot & 40.0 & 23.0 & --- & --- \\
BM25-all & 40.5 & 25.0 & --- & --- \\
BM25-geometry & 46.5 & 27.0 & --- & --- \\
Terminus 2 & --- & --- & 34.5 & --- \\
Terminus-KIRA & --- & --- & 43.1 & --- \\
mini-swe-agent & --- & --- & --- & 62.8 \\
\midrule
Meta-Harness & 46.0 & 29.0 & 44.8 & 63.6 \\
Ours & \textbf{62.0} & \textbf{30.5} & \textbf{50.0} & \textbf{66.0} \\
\bottomrule
\end{tabular*}
\end{table}

\section{Experiments}
\label{sec:results}

We first evaluate deployable performance using development-selected harnesses, then examine how the branches develop distinct mechanisms and how routing combines their capabilities.
We then assess the two adaptive components through search ablations and report retrospective test-best results as a reference for the quality of discovered candidates.

\subsection{Experimental Setup}
\label{sec:setup}

We evaluate on mathematical reasoning, terminal-based problem solving, and repository-level software engineering to assess whether the search procedure produces deployable gains across different tasks and action models.

\noindent\textbf{Mathematical reasoning.}
We follow the retrieval-augmented mathematical-reasoning setting of Meta-Harness \citep{lee2026metaharness}, in which a harness can retrieve worked solutions from a corpus of more than 500,000 problems to solve Olympiad-level questions.
We use 250 problems for development and 200 held-out IMO-level problems for testing.
% \yz{Please add citations to the official model reports or documentation for Gemini 3 Flash and Claude Sonnet 4.5}
We evaluate Gemini 3 Flash \citep{google2025gemini3flash} and Claude Sonnet 4.5 \citep{anthropic2025sonnet45} separately as action models and report held-out accuracy.

\noindent\textbf{Terminal-Bench 2.0.}
Terminal-Bench 2.0 evaluates agents on long-horizon tasks in realistic software environments \citep{merrill2026terminalbench}.
We randomly split the 88 tasks into 30 development tasks and 58 held-out test tasks.
All harnesses use Claude Sonnet 4.5 as the action model.
We evaluate task completion using the benchmark's task-specific checks and report the held-out pass rate.

\noindent\textbf{SWE-bench Lite.}
SWE-bench Lite pairs GitHub issues with repository snapshots and executable tests \citep{jimenez2024swebench}.
We randomly split its 300 tasks into 50 development tasks and 250 held-out test tasks.
Claude Sonnet 4.5 is used as the action model.
We evaluate generated patches using the benchmark tests and report the held-out issue-resolution rate (\%).

\begin{figure}[!t]
\centering
\includegraphics[width=\textwidth]{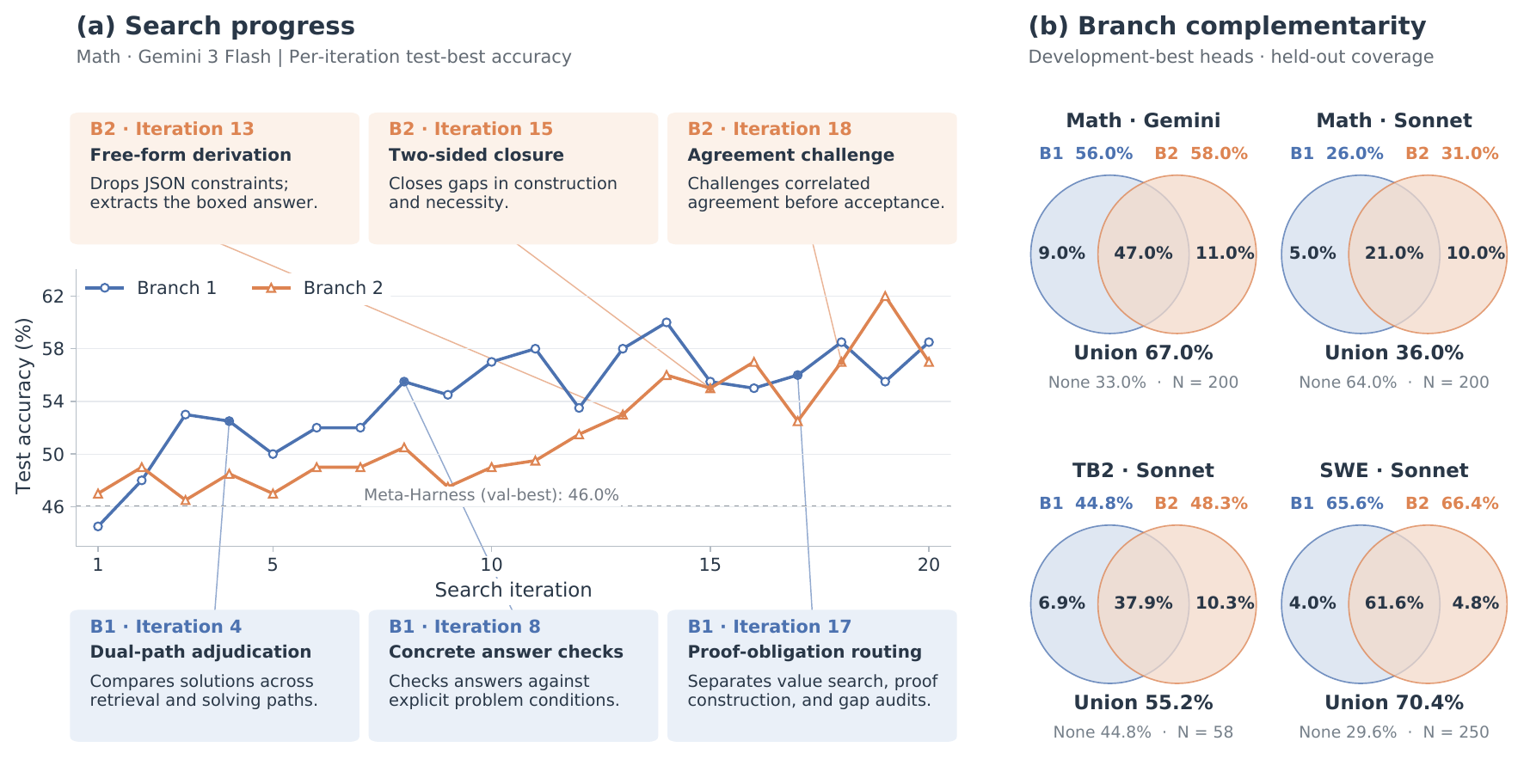}
\caption{Left: per-iteration test-best accuracy and discovered mechanisms on Math--Gemini over 20 iterations.
Right: shared and branch-exclusive coverage of development-selected heads in four settings.
All rates (\%) are computed over the full test sets.
Venn areas are schematic.}
\label{fig:math-gemini-trajectory-coverage}
\end{figure}

\noindent\textbf{Baselines.}
We compare against fixed harnesses and automated harness search.
Fixed baselines comprise no-memory and few-shot prompting, full-corpus and geometry-filtered BM25 retrieval for math \citep{robertson2009bm25}, Terminus 2 \citep{merrill2026terminalbench} and Terminus-KIRA \citep{terminuskira2026} for Terminal-Bench 2.0, and unmodified mini-swe-agent\footnote{\url{https://github.com/SWE-agent/mini-swe-agent}} \citep{yang2024sweagent} for SWE-bench Lite.
Meta-Harness \citep{lee2026metaharness} is our search-based baseline, evolving harnesses with a fixed development set and proposal guidance.

\noindent\textbf{Implementation details.}
Unless otherwise stated, we select harnesses using development performance and report their held-out test performance.
Meta-Harness selects one harness on the full development set, while ours routes among heads selected on each branch's final development subset.
% \yz{Could we condense this paragraph to one sentence stating that development-selected branch heads are routed at test time, with retrospective test-best results reported separately in Section~\ref{sec:test-reference}? The candidate counts and ranking details could be explained there.}
All settings generate one candidate harness per branch per iteration.
Throughout the experiments, we use $B=2$ branches, $N=20$ search iterations, frontier size $q=5$, and ownership margin $\Delta=2$.
The default update intervals are $C_p=1$ for development subsets and $C_s=5$ for proposal guidance, with ablations varying the update schedule or disabling individual components.
Appendix~\ref{app:token-accounting} reports task-solving and harness-authoring token usage across full search runs.

\subsection{Performance of Development-Selected Systems}
\label{sec:main-results}
\label{sec:deployment}

Table~\ref{tab:main-results} compares our routed system with fixed harnesses and Meta-Harness under development-based selection.
Our system surpasses the strongest fixed harness in every setting, with relative gains of 33.3\% and 13.0\% over BM25-geometry for the two math models, 16.0\% over Terminus-KIRA on Terminal-Bench 2.0, and 5.1\% over mini-swe-agent on SWE-bench Lite.
These improvements support the usefulness of harness evolution as a practical form of recursive self-improvement.
Compared with Meta-Harness, our system achieves further relative gains of 34.8\% on Math--Gemini 3 Flash, 5.2\% on Math--Claude Sonnet 4.5, 11.6\% on Terminal-Bench 2.0, and 3.8\% on SWE-bench Lite.
The gains over Meta-Harness are largest on Math--Gemini and Terminal-Bench 2.0, but extend across both mathematical reasoning and agentic coding, supporting applicability across the evaluated tasks and action models.

\subsection{How Branches Develop Different Search Trajectories}
\label{sec:search-analysis}

We use Math--Gemini to examine how development-subset and proposal-guidance updates shape distinct search trajectories, then assess whether the resulting heads exhibit complementary capabilities across all four settings.
The left panel of Figure~\ref{fig:math-gemini-trajectory-coverage} highlights the mechanisms discovered by each branch on Math--Gemini.
Branch~1 develops concrete answer checks and proof-obligation routing, focusing on verifying proposed answers and identifying the type of argument a problem requires.
Branch~2 explores free-form derivation, two-sided closure, and agreement challenges, focusing on constructing solutions and checking for shared errors when solvers agree.
% \yz{Could we support these descriptions with concrete case studies showing how each branch's harness handles a problem? An example connecting a development-subset or guidance update to a subsequent harness modification would also help substantiate how adaptation shapes the search trajectories.}
The trajectories thus develop different approaches to answer verification and solution construction.

\begin{table}[!t]
\caption{Held-out performance (\%) of development-selected branch heads and the routed system.}
\label{tab:router-ablation}
\centering
\small
\setlength{\tabcolsep}{4pt}
\renewcommand{\arraystretch}{1.12}
\setlength{\tabcolsep}{6pt}
\begin{tabular*}{\textwidth}{@{\extracolsep{\fill}}ccccc@{}}
\toprule
& \multicolumn{2}{c}{\textbf{Math}} & \textbf{Terminal-Bench 2.0} & \textbf{SWE-bench Lite} \\
\cmidrule(lr){2-3}\cmidrule(lr){4-4}\cmidrule(l){5-5}
& \shortstack{Gemini 3\\Flash} & \shortstack{Claude Sonnet\\4.5} & \shortstack{Claude Sonnet\\4.5} & \shortstack{Claude Sonnet\\4.5} \\
\midrule
Head 1 only & 56.0 & 26.0 & 44.8 & 65.6 \\
Head 2 only & 58.0 & \textbf{31.0} & 48.3 & \textbf{66.4} \\
Router (ours) & \textbf{62.0} & 30.5 & \textbf{50.0} & 66.0 \\
\bottomrule
\end{tabular*}

\vspace{14pt}
\centering
\caption{Math--Gemini held-out accuracy (\%) with default routing: (a) development-update frequency and (b) development pruning and optimizer updates.}
\label{tab:search-ablation}
\small
\renewcommand{\arraystretch}{1.15}
\begin{minipage}[t]{0.43\textwidth}
\centering
\textbf{(a) Development-update frequency}\par\smallskip
\begin{tabular*}{\linewidth}{@{\extracolsep{\fill}}cc@{}}
\toprule
Update interval & Accuracy \\
\midrule
Every 5 iterations & 53.0 \\
Every 3 iterations & 56.0 \\
Every iteration (ours) & \textbf{62.0} \\
\bottomrule
\end{tabular*}
\end{minipage}\hfill
\begin{minipage}[t]{0.53\textwidth}
\centering
\textbf{(b) Component settings}\par\smallskip
\begin{tabular*}{\linewidth}{@{\extracolsep{\fill}}ccc@{}}
\toprule
\shortstack{Optimizer updates $\rightarrow$\\Dev prune $\downarrow$} & \raisebox{0.5\baselineskip}{Off} & \raisebox{0.5\baselineskip}{On} \\[2.33pt]
\midrule
Off & 51.0 & 50.0 \\
On & 54.0 & \textbf{62.0} \\
\bottomrule
\end{tabular*}
\end{minipage}

\end{table}

We further examine how subset changes reshape the search trajectory, with subset histories in Appendix~\ref{app:subset-histories}.
For example, at iteration~13 on Math--Gemini, Branch~2 removes four cases, reducing its subset from 44 to 40 problems.
In the same iteration, a free-form derivation harness replaces the geometry-aware head, with an expanded analysis in Appendix~\ref{app:head-reselection}.
Subsequent harnesses extend the new head with progressive derivation and two-sided closure.

Evolving proposal policies also shape the search trajectories.
For example, we compare the branches' proposal guidance after iteration~15, $S_1^{15}$ and $S_2^{15}$.
Branch~1 prioritizes extracting concrete information from retrieved solutions, while Branch~2 emphasizes restructuring derivations after free-form derivation, progressive refinement, and two-sided closure become successive heads.
These priorities direct further exploration toward improving the use of external evidence in one branch and the construction of solutions in the other.
Appendix~\ref{app:gradient-examples} gives concrete examples of how proposer $P$ updates branch-specific guidance $S_b^t$.

A Math--Gemini development case illustrates how Branch~1's targeted answer checks help it solve a case missed by Branch~2.
The case asks for all real-coefficient polynomials that map positive integers consisting entirely of ones to integers of the same form.
Finding a valid family is insufficient because the answer must include every such polynomial.
Within Branch~1's harness, two candidate answers agree on the polynomial form but disagree on whether a parameter can be negative.
Rather than accepting the narrower range, the harness tests an excluded example, $f(x)=(9x^2+2x-1)/10$, which maps a repunit with $k$ digits to one with $2k-1$ digits.
This counterexample exposes the unnecessary restriction and leads the harness to retain the broader, correct family.
This recovery illustrates how Branch~1's concrete answer checks, shown in the left panel of Figure~\ref{fig:math-gemini-trajectory-coverage}, detect valid solutions omitted by a candidate answer.

Beyond this Math--Gemini case study, the right panel of Figure~\ref{fig:math-gemini-trajectory-coverage} shows that the development-selected heads solve complementary held-out cases across all four settings.
Each branch solves cases missed by the other, raising combined coverage by 4.0--9.0\% in absolute terms over the stronger individual head.
This oracle upper bound can only be reached if the router selects a successful expert whenever one is available.
% \yz{Could we characterize the task types or reasoning requirements among each branch's uniquely solved cases and relate them to its harness design? This would clarify what the complementary capabilities are beyond the coverage numbers. We could summarize the main finding here and place the detailed breakdown and representative examples in the appendix.}

\begin{table}[t]
\centering
\caption{Test-best performance (\%) among ten harnesses per method: the top five by development score per branch for ours and the top ten for Meta-Harness.}
\label{tab:test-best-results}
\small
\setlength{\tabcolsep}{7pt}
\renewcommand{\arraystretch}{1.12}
\begin{tabular*}{\textwidth}{@{\extracolsep{\fill}}ccccc@{}}
\toprule
\textbf{Method} & \multicolumn{2}{c}{\textbf{Math}} & \textbf{Terminal-Bench 2.0} & \textbf{SWE-bench Lite} \\
\cmidrule(lr){2-3}\cmidrule(lr){4-4}\cmidrule(l){5-5}
& \shortstack{Gemini 3\\Flash} & \shortstack{Claude Sonnet\\4.5} & \shortstack{Claude Sonnet\\4.5} & \shortstack{Claude Sonnet\\4.5} \\
\midrule
Meta-Harness & 49.0 & \textbf{32.5} & 46.6 & 63.6 \\
Ours & \textbf{58.0} & \textbf{32.5} & \textbf{50.0} & \textbf{66.4} \\
\bottomrule
\end{tabular*}
\end{table}

\subsection{Router Analysis}
\label{sec:router-analysis}

Table~\ref{tab:router-ablation} compares routing with the stronger of the two development-selected branch heads.
The router exceeds this reference by 4.0\% on Math--Gemini and 1.7\% on Terminal-Bench 2.0 in absolute terms, while falling short by one test case on Math--Sonnet and SWE-bench Lite.
The router thus approaches or exceeds the stronger head without test feedback.
Routing nevertheless remains 4.4--5.5\% below combined coverage in absolute terms, leaving some complementary capabilities unrealized.
For example, the Math--Gemini router succeeds on 12/18 (66.7\%) Branch~1-only cases and 18/22 (81.8\%) Branch~2-only cases, recovering 75.0\% of their exclusive successes.
The remaining 10 missed cases explain its 5.0\% gap to combined coverage.
% \yz{Could we analyze routing accuracy specifically on cases solved by exactly one head, separately for each branch? This would show whether the router recognizes both branches' strengths or disproportionately favors one.}

To assess the contributions of router context and prompt optimization, we compare three input configurations: expert source code alone, code with head-exclusive development examples, and code with both examples and expert outputs.
Each configuration is evaluated with and without GEPA.
Without GEPA, adding expert outputs to these examples raises Math--Gemini accuracy from 57.0\% to 59.5\%, but lowers Terminal-Bench 2.0 accuracy from 48.3\% to 46.6\%.
Additional context therefore offers setting-dependent benefits rather than a consistent improvement.
% \yz{Could we inspect cases where adding examples or expert outputs changes the routing decision, distinguishing corrections from newly introduced errors? A few examples could clarify when this context reveals relevant capabilities and when it encourages misleading matches to development cases. Treat these explanations as qualitative observations unless supported by a broader breakdown.}
GEPA provides modest average gains across the three input configurations, with the largest benefit from combining examples and expert outputs.
Appendix~\ref{app:router-context} reports the full ablation results.

\subsection{Method Ablation Studies}
\label{sec:ablations}

\noindent\textbf{More frequent development updates improve performance.}
Table~\ref{tab:search-ablation}a tests how quickly branch objectives should respond to emerging coverage differences by varying the development-update interval.
We update the subsets every 1, 3, or 5 iterations, obtaining 62.0\%, 56.0\%, and 53.0\% accuracy on Math--Gemini, respectively.
Between development updates, each branch continues proposing harnesses against its unchanged subset.
Longer intervals therefore allow more proposals before newly observed differences between branches affect the search objective, potentially prolonging exploration of problems that an earlier update would remove.
The trend is consistent with timely specialization benefiting the routed system, although these aggregate scores do not establish whether update frequency changes the diversity of discovered mechanisms.

\noindent\textbf{Combining development updates and proposal adaptation performs best.}
Table~\ref{tab:search-ablation}b compares all four combinations of development pruning and optimizer updates.
Here, optimizer updates revise branch-local proposal guidance, rather than model weights or optimizer code.
Disabling them keeps this guidance constant throughout search.
Without development pruning, both branches retain the same fixed development set, yielding two independent Meta-Harness searches with either fixed or evolving proposal policies.
All four conditions deploy the default router over development-selected branch heads.
Enabling both components achieves 62.0\% accuracy, compared with 51.0\% with neither, 50.0\% with optimizer updates alone, and 54.0\% with pruning alone.
Proposal adaptation alone does not improve on independent search, but increases accuracy when development pruning is enabled.
These results suggest that different development subsets make proposal adaptation more useful by giving each branch distinct search challenges.

\subsection{Test Performance Analysis}
\label{sec:test-reference}

To assess the development--test selection gap, we evaluate 10 harnesses per method: the top 5 from each of our branches ranked on its final development subset and Meta-Harness's top 10 on the full development set.
Table~\ref{tab:test-best-results} reports the highest test accuracy within each pool.
For Meta-Harness, test-based selection increases Math--Gemini accuracy from 46.0\% to 49.0\% and Math--Sonnet accuracy from 29.0\% to 32.5\%.
These gaps reflect imperfect development rankings and optimistic test-based selection, complementing concerns about evaluation on search tasks~\citep{wang2026rethinking}.

Our test-best harness outperforms Meta-Harness's test-best harness in three settings and matches it on Math--Sonnet.
These results support branching as a more effective search strategy for discovering high-performing harnesses under the evaluated configurations.
Moreover, on Math--Gemini, our router reaches 62.0\% accuracy against 58.0\% for the test-best individual harness in the pool, despite relying solely on development data for selection.
This illustrates why the best individual harness's test score need not bound the performance achievable by routing among complementary experts.
Appendix~\ref{app:harness-analysis} details the test-best harness from each setting.

\section{Conclusion}

We introduced branching search in which development-case assignments evolve alongside branch-local proposer guidance.
Comparative frontier coverage changes the objective used to select each branch's parent, and separate skill documents preserve local priorities for subsequent proposals.
A recorded head replacement demonstrates how this mechanism redirects search toward a different family of harnesses.
Across four settings, development-selected heads with routing outperform development-selected Meta-Harness without test-based harness selection.

The evaluation establishes gains under the configured search procedures, while unequal total token usage and limited repeated evaluations constrain conclusions about efficiency and reliability.
Routing also falls slightly below the stronger individual head in two settings.
The component ablations support the combined design, but do not establish that each observed trajectory change caused its subsequent test gain.
The results support adapting search objectives and branch-local proposal guidance together, then using development-informed routing to deploy the complementary harnesses.

\subsection*{AI use statement}

In this work, we used generative AI tools to assist in implementing the experimental system, providing feedback on experimental design, analyzing search trajectories and discovered harness mechanisms, and interpreting results.
We did not use generative AI tools to develop the core methodology or propose research hypotheses.
Additionally, we used these tools to draft and revise manuscript text, organize the paper, identify and summarize relevant literature, and create or modify scientific figures and supporting code.
The LLM-based generation of harnesses, adaptation of proposal guidance, and expert routing are components of the research method described in the paper.
We take responsibility for the final content of this work, including text, claims, and artifacts produced with the aid of generative AI.

\bibliography{iclr2027_conference}

\begin{thebibliography}{39}
\providecommand{\natexlab}[1]{#1}
\providecommand{\url}[1]{\texttt{#1}}
\expandafter\ifx\csname urlstyle\endcsname\relax
  \providecommand{\doi}[1]{doi: #1}\else
  \providecommand{\doi}{doi: \begingroup \urlstyle{rm}\Url}\fi

\bibitem[Agrawal et~al.(2026)Agrawal, Tan, Soylu, Ziems, Khare, Opsahl-Ong,
  Singhvi, Shandilya, Ryan, Jiang, Potts, Sen, Dimakis, Stoica, Klein, Zaharia,
  and Khattab]{agrawal2025gepa}
Lakshya~A. Agrawal, Shangyin Tan, Dilara Soylu, Noah Ziems, Rishi Khare, Krista
  Opsahl-Ong, Arnav Singhvi, Herumb Shandilya, Michael~J. Ryan, Meng Jiang,
  Christopher Potts, Koushik Sen, Alexandros~G. Dimakis, Ion Stoica, Dan Klein,
  Matei Zaharia, and Omar Khattab.
\newblock {GEPA}: Reflective prompt evolution can outperform reinforcement
  learning.
\newblock In \emph{International Conference on Learning Representations}, 2026.

\bibitem[{Anthropic}(2025)]{anthropic2025sonnet45}
{Anthropic}.
\newblock {Claude Sonnet 4.5 System Card}.
\newblock Technical report, Anthropic, September 2025.
\newblock URL
  \url{https://assets.anthropic.com/m/12f214efcc2f457a/original/Claude-Sonnet-4-5-System-Card.pdf}.

\bibitem[{Anthropic}(2026)]{anthropic2026opus46}
{Anthropic}.
\newblock {Claude Opus 4.6 System Card}.
\newblock Technical report, Anthropic, February 2026.
\newblock URL
  \url{https://www-cdn.anthropic.com/0dd865075ad3132672ee0ab40b05a53f14cf5288.pdf}.

\bibitem[Chen et~al.(2023)Chen, Zaharia, and Zou]{chen2023frugalgpt}
Lingjiao Chen, Matei Zaharia, and James Zou.
\newblock {FrugalGPT}: How to use large language models while reducing cost and
  improving performance.
\newblock \emph{arXiv preprint arXiv:2305.05176}, 2023.

\bibitem[Chen et~al.(2026{\natexlab{a}})Chen, Lv, Zhang, Chang, and
  Zhou]{chen2026harnessforge}
Mingju Chen, Can Lv, Guibin Zhang, Heng Chang, and Shiji Zhou.
\newblock Harnessforge: Joint harness and policy evolution for adaptive agent
  systems.
\newblock \emph{arXiv preprint arXiv:2606.01779}, 2026{\natexlab{a}}.

\bibitem[Chen et~al.(2026{\natexlab{b}})Chen, Xiao, Zhu, Yuan, Zhang, and
  Wang]{chen2026coharness}
Zhengyu Chen, Teng Xiao, Huaisheng Zhu, Yige Yuan, Luan Zhang, and Jingang
  Wang.
\newblock Co-harness: Co-evolving harnesses and model weights for llm agents.
\newblock \emph{arXiv preprint arXiv:2607.22688}, 2026{\natexlab{b}}.

\bibitem[Doshi(2025)]{google2025gemini3flash}
Tulsee Doshi.
\newblock {Gemini 3 Flash}: Frontier intelligence built for speed.
\newblock Google Blog, December 2025.
\newblock URL
  \url{https://blog.google/products-and-platforms/products/gemini/gemini-3-flash/}.

\bibitem[Feng et~al.(2026)Feng, Zeng, Grover, Qiu, Xia, Zhang, Wang, Chen, Fu,
  Liu, et~al.]{feng2026dag}
Jiarui Feng, Hanqing Zeng, Karish Grover, Ruizhong Qiu, Yinglong Xia, Qiang
  Zhang, Qifan Wang, Ren Chen, Dongqi Fu, Jiayi Liu, et~al.
\newblock Dag-moe: From simple mixture to structural aggregation in
  mixture-of-experts.
\newblock \emph{arXiv preprint arXiv:2606.01062}, 2026.

\bibitem[Guo et~al.(2026)Guo, Shi, Chen, Xu, Wang, Zhang, Ni, Zhu, and
  Di]{guo2026drevo}
Hanghui Guo, Weijie Shi, Zhangze Chen, Shengxiang Xu, Yishu Wang, Yimei Zhang,
  Wangze Ni, Jia Zhu, and Shimin Di.
\newblock {DREvo}: Distilling recalibrated historical experience for harness
  self-evolution.
\newblock \emph{arXiv preprint arXiv:2607.26722}, 2026.

\bibitem[Hao et~al.(2026)Hao, Long, and Zhao]{hao2026self}
Guangya Hao, Yunbo Long, and Zhuokai Zhao.
\newblock Self-evolving multi-agent systems via decentralized memory.
\newblock \emph{arXiv preprint arXiv:2605.22721}, 2026.

\bibitem[Hu et~al.(2024)Hu, Lu, and Clune]{hu2024adas}
Shengran Hu, Cong Lu, and Jeff Clune.
\newblock Automated design of agentic systems.
\newblock \emph{arXiv preprint arXiv:2408.08435}, 2024.

\bibitem[Jacobs et~al.(1991)Jacobs, Jordan, Nowlan, and
  Hinton]{jacobs1991mixtures}
Robert~A. Jacobs, Michael~I. Jordan, Steven~J. Nowlan, and Geoffrey~E. Hinton.
\newblock Adaptive mixtures of local experts.
\newblock \emph{Neural Computation}, 3\penalty0 (1):\penalty0 79--87, 1991.

\bibitem[Jimenez et~al.(2024)Jimenez, Yang, Wettig, Yao, Pei, Press, and
  Narasimhan]{jimenez2024swebench}
Carlos~E. Jimenez, John Yang, Alexander Wettig, Shunyu Yao, Kexin Pei, Ofir
  Press, and Karthik Narasimhan.
\newblock {SWE-bench}: Can language models resolve real-world {GitHub} issues?
\newblock In \emph{International Conference on Learning Representations}, 2024.

\bibitem[Karten et~al.(2026)Karten, Zhang, Upaa, Feng, Li, Shi, Jin, and
  Vodrahalli]{karten2026continualharness}
Seth Karten, Joel Zhang, Tersoo Upaa, Jr., Ruirong Feng, Wenzhe Li, Chengshuai
  Shi, Chi Jin, and Kiran Vodrahalli.
\newblock Continual harness: Online adaptation for self-improving foundation
  agents.
\newblock \emph{arXiv preprint arXiv:2605.09998}, 2026.

\bibitem[Khattab et~al.(2023)Khattab, Singhvi, Maheshwari, Zhang, Santhanam,
  Vardhamanan, Haq, Sharma, Joshi, Moazam, Miller, Zaharia, and
  Potts]{khattab2024dspy}
Omar Khattab, Arnav Singhvi, Paridhi Maheshwari, Zhiyuan Zhang, Keshav
  Santhanam, Sri Vardhamanan, Saiful Haq, Ashutosh Sharma, Thomas~T. Joshi,
  Hanna Moazam, Heather Miller, Matei Zaharia, and Christopher Potts.
\newblock {DSPy}: Compiling declarative language model calls into
  self-improving pipelines.
\newblock \emph{arXiv preprint arXiv:2310.03714}, 2023.

\bibitem[{KRAFTON AI} \& {Ludo Robotics}(2026){KRAFTON AI} and {Ludo
  Robotics}]{terminuskira2026}
{KRAFTON AI} and {Ludo Robotics}.
\newblock {Terminus-KIRA}: Boosting frontier model performance on
  {Terminal-Bench} with minimal harness, 2026.
\newblock URL \url{https://github.com/krafton-ai/KIRA}.

\bibitem[Lee et~al.(2026)Lee, Nair, Zhang, Lee, Khattab, and
  Finn]{lee2026metaharness}
Yoonho Lee, Roshen Nair, Qizheng Zhang, Kangwook Lee, Omar Khattab, and Chelsea
  Finn.
\newblock Meta-harness: End-to-end optimization of model harnesses.
\newblock \emph{arXiv preprint arXiv:2603.28052}, 2026.

\bibitem[Lehman \& Stanley(2011)Lehman and Stanley]{lehman2011novelty}
Joel Lehman and Kenneth~O. Stanley.
\newblock Abandoning objectives: Evolution through the search for novelty
  alone.
\newblock \emph{Evolutionary Computation}, 19\penalty0 (2):\penalty0 189--223,
  2011.

\bibitem[Lewis et~al.(2020)Lewis, Perez, Piktus, Petroni, Karpukhin, Goyal,
  K{\"u}ttler, Lewis, Yih, Rockt{\"a}schel, Riedel, and Kiela]{lewis2020rag}
Patrick Lewis, Ethan Perez, Aleksandra Piktus, Fabio Petroni, Vladimir
  Karpukhin, Naman Goyal, Heinrich K{\"u}ttler, Mike Lewis, Wen-tau Yih, Tim
  Rockt{\"a}schel, Sebastian Riedel, and Douwe Kiela.
\newblock Retrieval-augmented generation for knowledge-intensive {NLP} tasks.
\newblock In \emph{Advances in Neural Information Processing Systems}, 2020.

\bibitem[Luo et~al.(2026)Luo, Jiang, Zou, Huang, Yan, Li, Yue, Li, Chen, Zhao,
  Liu, Cui, Shen, and Li]{luo2026autodesign}
Yaxin Luo, Haobin Jiang, Jialv Zou, Xu~Huang, Wenhao Yan, Haodong Li, Zhengrong
  Yue, Jing Li, Xiaofu Chen, Xiaohan Zhao, Jiacheng Liu, Jiacheng Cui, Zhiqiang
  Shen, and Xiaotong Li.
\newblock Autodesign: Meta-harness optimization for long-horizon agentic
  design.
\newblock \emph{arXiv preprint arXiv:2608.13560}, 2026.

\bibitem[Merrill et~al.(2026)Merrill, Shaw, Carlini, Li, Raj, Bercovich, Shi,
  Shin, Walshe, Buchanan, et~al.]{merrill2026terminalbench}
Mike~A. Merrill, Alexander~G. Shaw, Nicholas Carlini, Boxuan Li, Harsh Raj,
  Ivan Bercovich, Lin Shi, Jeong~Yeon Shin, Thomas Walshe, E.~Kelly Buchanan,
  et~al.
\newblock Terminal-bench: Benchmarking agents on hard, realistic tasks in
  command line interfaces.
\newblock \emph{arXiv preprint arXiv:2601.11868}, 2026.

\bibitem[Mouret \& Clune(2015)Mouret and Clune]{mouret2015mapelites}
Jean-Baptiste Mouret and Jeff Clune.
\newblock Illuminating search spaces by mapping elites.
\newblock \emph{arXiv preprint arXiv:1504.04909}, 2015.

\bibitem[Novikov et~al.(2025)Novikov, V{\~u}, Eisenberger, Dupont, Huang,
  Wagner, Shirobokov, Kozlovskii, Ruiz, Mehrabian, Kumar, See, Chaudhuri,
  Holland, Davies, Nowozin, Kohli, and Balog]{novikov2025alphaevolve}
Alexander Novikov, Ng{\^a}n V{\~u}, Marvin Eisenberger, Emilien Dupont, Po-Sen
  Huang, Adam~Zsolt Wagner, Sergey Shirobokov, Borislav Kozlovskii, Francisco
  J.~R. Ruiz, Abbas Mehrabian, M.~Pawan Kumar, Abigail See, Swarat Chaudhuri,
  George Holland, Alex Davies, Sebastian Nowozin, Pushmeet Kohli, and Matej
  Balog.
\newblock {AlphaEvolve}: A coding agent for scientific and algorithmic
  discovery.
\newblock \emph{arXiv preprint arXiv:2506.13131}, 2025.

\bibitem[Ong et~al.(2024)Ong, Almahairi, Wu, Chiang, Wu, Gonzalez, Kadous, and
  Stoica]{ong2024routellm}
Isaac Ong, Amjad Almahairi, Vincent Wu, Wei-Lin Chiang, Tianhao Wu, Joseph~E.
  Gonzalez, M.~Waleed Kadous, and Ion Stoica.
\newblock {RouteLLM}: Learning to route llms with preference data.
\newblock \emph{arXiv preprint arXiv:2406.18665}, 2024.

\bibitem[Pryzant et~al.(2023)Pryzant, Iter, Li, Lee, Zhu, and
  Zeng]{pryzant2023protegi}
Reid Pryzant, Dan Iter, Jerry Li, Yin~Tat Lee, Chenguang Zhu, and Michael Zeng.
\newblock Automatic prompt optimization with ``gradient descent'' and beam
  search.
\newblock In \emph{Proceedings of the 2023 Conference on Empirical Methods in
  Natural Language Processing}, 2023.

\bibitem[Robertson \& Zaragoza(2009)Robertson and Zaragoza]{robertson2009bm25}
Stephen Robertson and Hugo Zaragoza.
\newblock The probabilistic relevance framework: {BM25} and beyond.
\newblock \emph{Foundations and Trends in Information Retrieval}, 3\penalty0
  (4):\penalty0 333--389, 2009.
\newblock \doi{10.1561/1500000019}.

\bibitem[Sengupta \& Wang(2026)Sengupta and Wang]{sengupta2026harbor}
Biswa Sengupta and Jinhua Wang.
\newblock {HARBOR}: Automated harness optimization.
\newblock \emph{arXiv preprint arXiv:2604.20938}, 2026.

\bibitem[Shinn et~al.(2023)Shinn, Cassano, Gopinath, Narasimhan, and
  Yao]{shinn2023reflexion}
Noah Shinn, Federico Cassano, Ashwin Gopinath, Karthik Narasimhan, and Shunyu
  Yao.
\newblock Reflexion: Language agents with verbal reinforcement learning.
\newblock In \emph{Advances in Neural Information Processing Systems}, 2023.

\bibitem[Wang et~al.(2026)Wang, Zhu, Hu, Yuan, Chen, Senthil, Hajishirzi,
  Tsvetkov, Dasigi, and Xiao]{wang2026rethinking}
Yike Wang, Huaisheng Zhu, Zhengyu Hu, Yige Yuan, Zhengyu Chen, Shakti Senthil,
  Hannaneh Hajishirzi, Yulia Tsvetkov, Pradeep Dasigi, and Teng Xiao.
\newblock Rethinking the evaluation of harness evolution for agents.
\newblock \emph{arXiv preprint arXiv:2607.12227}, 2026.

\bibitem[Xiong et~al.(2026)Xiong, Zhou, Zeng, Chen, Huang, Bi, Zhang, and
  Zhao]{xiong2026token}
Nuoya Xiong, Yuhang Zhou, Hanqing Zeng, Zhaorun Chen, Furong Huang, Shuchao Bi,
  Lizhu Zhang, and Zhuokai Zhao.
\newblock Token-level llm collaboration via fusionroute.
\newblock \emph{arXiv preprint arXiv:2601.05106}, 2026.

\bibitem[Yang et~al.(2024{\natexlab{a}})Yang, Wang, Lu, Liu, Le, Zhou, and
  Chen]{yang2024opro}
Chengrun Yang, Xuezhi Wang, Yifeng Lu, Hanxiao Liu, Quoc~V. Le, Denny Zhou, and
  Xinyun Chen.
\newblock Large language models as optimizers.
\newblock In \emph{International Conference on Learning Representations},
  2024{\natexlab{a}}.

\bibitem[Yang et~al.(2024{\natexlab{b}})Yang, Jimenez, Wettig, Lieret, Yao,
  Narasimhan, and Press]{yang2024sweagent}
John Yang, Carlos~E. Jimenez, Alexander Wettig, Kilian Lieret, Shunyu Yao,
  Karthik Narasimhan, and Ofir Press.
\newblock {SWE-agent}: Agent-computer interfaces enable automated software
  engineering.
\newblock In \emph{Advances in Neural Information Processing Systems},
  2024{\natexlab{b}}.
\newblock URL \url{https://arxiv.org/abs/2405.15793}.

\bibitem[Yang et~al.(2026)Yang, Gong, Huang, Yang, Zhou, Huang, Li, Gao, Dai,
  Liu, Qiu, Yang, Chen, Yang, and Luo]{yang2026skillopt}
Yifan Yang, Ziyang Gong, Weiquan Huang, Qihao Yang, Ziwei Zhou, Zisu Huang, Yan
  Li, Xuemei Gao, Qi~Dai, Bei Liu, Kai Qiu, Yuqing Yang, Dongdong Chen, Xue
  Yang, and Chong Luo.
\newblock {SkillOpt}: Executive strategy for self-evolving agent skills.
\newblock \emph{arXiv preprint arXiv:2605.23904}, 2026.

\bibitem[Yao et~al.(2023)Yao, Zhao, Yu, Du, Shafran, Narasimhan, and
  Cao]{yao2023react}
Shunyu Yao, Jeffrey Zhao, Dian Yu, Nan Du, Izhak Shafran, Karthik Narasimhan,
  and Yuan Cao.
\newblock {ReAct}: Synergizing reasoning and acting in language models.
\newblock In \emph{International Conference on Learning Representations}, 2023.

\bibitem[Yuan et~al.(2026)Yuan, Wang, Wang, Sun, Wang, and
  Li]{yuan2026agenticgeo}
Jiaqi Yuan, Jialu Wang, Zihan Wang, Qingyun Sun, Ruijie Wang, and Jianxin Li.
\newblock {AgenticGEO}: A self-evolving agentic system for generative engine
  optimization.
\newblock \emph{arXiv preprint arXiv:2603.20213}, 2026.

\bibitem[Yuksekgonul et~al.(2024)Yuksekgonul, Bianchi, Boen, Liu, Huang,
  Guestrin, and Zou]{yuksekgonul2024textgrad}
Mert Yuksekgonul, Federico Bianchi, Joseph Boen, Sheng Liu, Zhi Huang, Carlos
  Guestrin, and James Zou.
\newblock {TextGrad}: Automatic ``differentiation'' via text.
\newblock \emph{arXiv preprint arXiv:2406.07496}, 2024.

\bibitem[Zeng et~al.(2025)Zeng, Xia, Zhao, Jiang, Zhang, Liu, Zhang, Zhang,
  Fan, and Zhang]{zeng2026s}
Hanqing Zeng, Yinglong Xia, Zhuokai Zhao, Chuan Jiang, Qiang Zhang, Jiayi Liu,
  Qunshu Zhang, Lizhu Zhang, Xiangjun Fan, and Benyu Zhang.
\newblock {S'MoRE}: Structural mixture of residual experts for
  parameter-efficient {LLM} fine-tuning.
\newblock In \emph{Advances in Neural Information Processing Systems},
  volume~38, pp.\  81441--81478, 2025.

\bibitem[Zhang et~al.(2026)Zhang, Zhang, Li, Zhang, Chen, Zhang, Bai, and
  Hu]{zhang2026selfharness}
Hangfan Zhang, Shao Zhang, Kangcong Li, Chen Zhang, Yang Chen, Yiqun Zhang, Lei
  Bai, and Shuyue Hu.
\newblock Self-harness: Harnesses that improve themselves.
\newblock \emph{arXiv preprint arXiv:2606.09498}, 2026.

\bibitem[Zhang et~al.(2024)Zhang, Xiang, Yu, Teng, Chen, Chen, Zhuge, Cheng,
  Hong, Wang, Zheng, Liu, Luo, and Wu]{zhang2024aflow}
Jiayi Zhang, Jinyu Xiang, Zhaoyang Yu, Fengwei Teng, Xionghui Chen, Jiaqi Chen,
  Mingchen Zhuge, Xin Cheng, Sirui Hong, Jinlin Wang, Bingnan Zheng, Bang Liu,
  Yuyu Luo, and Chenglin Wu.
\newblock {AFlow}: Automating agentic workflow generation.
\newblock \emph{arXiv preprint arXiv:2410.10762}, 2024.

\end{thebibliography}
\bibliographystyle{iclr2027_conference}

\appendix
\setcounter{topnumber}{3}
\setcounter{bottomnumber}{2}
\setcounter{totalnumber}{5}
\section{Token Accounting}
\label{app:token-accounting}

We account for full-run token usage by separating task solving from harness authoring.
Across the three benchmark settings, task solving uses Claude Sonnet 4.5 as the action model $M$, while harness authoring uses Claude Opus 4.6 as the proposer $P$.
Task-solving usage sums input and output tokens across recorded evaluation calls, including cached input once.
Harness-authoring usage sums uncached input, cache-read, cache-creation, and output tokens across proposer sessions.
Total usage combines both components.
Table~\ref{tab:token-accounting} reports these totals for each benchmark.
The initial development pools contain 250 cases for math, 30 for Terminal-Bench 2.0, and 50 for SWE-bench Lite.
Despite using $B=2$ branches, ours consumes $1.21\times$, $1.71\times$, and $1.50\times$ as many task-solving tokens as Meta-Harness on these benchmarks, respectively.
Development-set pruning helps limit this overhead by reducing the number of cases evaluated for subsequent candidates.

\begin{table}[!htbp]
\centering
\caption{Full-run token usage (millions) for Meta-Harness and ours across three benchmarks with Claude Sonnet 4.5 as the action model and Claude Opus 4.6 as the proposer.
Task-solving and harness-authoring tokens sum to the total, with cached tokens counted once.}
\label{tab:token-accounting}
\small
\setlength{\tabcolsep}{3pt}
\renewcommand{\arraystretch}{1.10}
\begin{tabular*}{\textwidth}{@{\extracolsep{\fill}}lrrrrrr@{}}
\toprule
& \multicolumn{3}{c}{\textbf{Meta-Harness}} & \multicolumn{3}{c}{\textbf{Ours}} \\
\cmidrule(lr){2-4}\cmidrule(l){5-7}
Benchmark & \shortstack{Task\\solving} & \shortstack{Harness\\authoring} & Total & \shortstack{Task\\solving} & \shortstack{Harness\\authoring} & Total \\
\midrule
Math & 11.10 & 68.82 & 79.92 & 13.40 & 159.50 & 172.90 \\
Terminal-Bench 2.0 & 2,086.10 & 280.70 & 2,366.80 & 3,564.60 & 436.79 & 4,001.39 \\
SWE-bench Lite & 1,814.60 & 92.57 & 1,907.17 & 2,726.70 & 224.78 & 2,951.48 \\
\bottomrule
\end{tabular*}
\end{table}

\section{Development-Set Changes and Head Selection}
\label{app:development-ownership}

\subsection{Subset Histories Across Settings}
\label{app:subset-histories}

\begin{figure}[t]
\centering
\includegraphics[width=\textwidth]{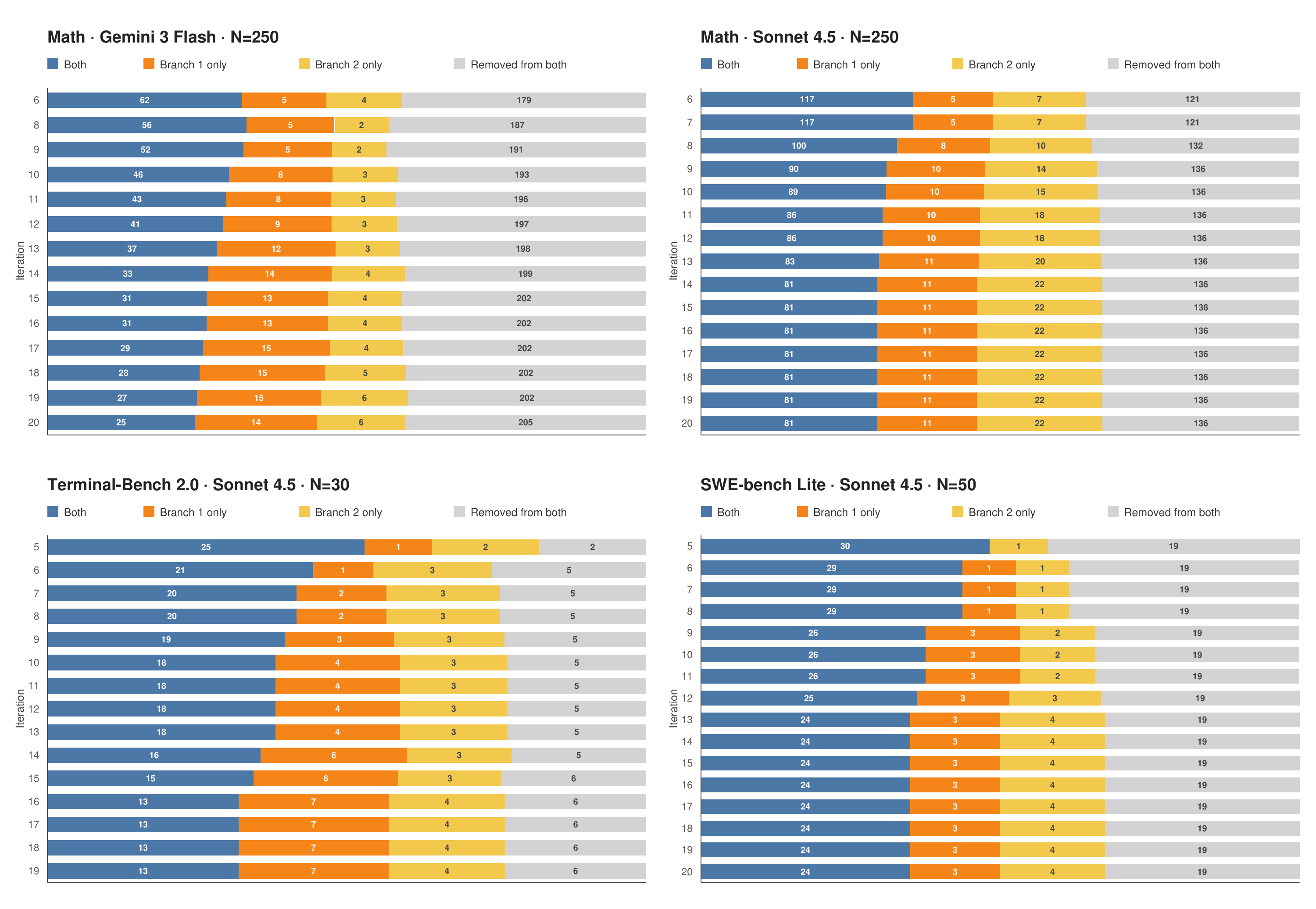}
\vspace{-0.6em}
\caption{Development-subset changes across iterations in four settings.
Filled segments show exact shared, branch-exclusive, and cumulatively removed counts, with widths scaled by $\log(1+n)$.
}
\label{fig:development-case-ownership}
\end{figure}

Figure~\ref{fig:development-case-ownership} tracks how development subsets shrink and diverge during search across the four settings.
Each branch's active subset comprises the shared cases and those retained exclusively by that branch.
Updates require $q=5$ candidate harnesses per branch and normally begin at iteration~5, while Math--Sonnet starts at iteration~6 because iteration~3 produced no candidate.

Across all four settings, the shared pool contracts while the number of branch-exclusive cases increases, giving the branches increasingly distinct evaluation sets.
On SWE-bench Lite, the number of cases removed from both branches remains at 19 between iterations~5 and~20, while shared cases decrease from 30 to 24 and branch-exclusive cases increase from 1 to 7.
Subset updates therefore promote specialization even when the total pool of retained cases stays unchanged.

\subsection{How Subset Changes Redirect Head Selection}
\label{app:head-reselection}

We expand the Math--Gemini Branch~2 transition described in Section~\ref{sec:search-analysis}.
Between iterations~8 and~12, its development subset contracts from 58 to 44 cases, while the geometry-aware dual-solver harness remains the head.
At iteration~13, the subset loses four more cases and the newly proposed \texttt{freeform\_derivation\_dual} becomes the head.
This transition changes the design that subsequent proposals use as their starting point.

The new harness preserves the previous head's retrieval paths, geometry-specific corpus selection, and conditional adjudication, but changes how the solvers present their reasoning.
Instead of writing derivations inside a JSON field, they produce ordinary mathematical prose and LaTeX, with the final answer extracted separately.
Its design rationale connects this change to the remaining development failures, noting that earlier attempts to impose additional solution procedures had performed poorly.
The proposer consequently explores a less constrained derivation format while retaining the established retrieval structure.

At iteration~14, the proposal record identifies the free-form harness as the head, solving 8 of the 40 retained cases, followed by an earlier progressive-refinement harness solving 7.
The resulting \texttt{freeform\_progressive} explicitly combines the new head's free-form reasoning with the earlier harness's targeted retrieval.
It identifies a weak step in the initial derivation, retrieves relevant examples, and derives the solution again without the JSON constraint.
At iteration~15, this combined harness is recorded as the head and the starting point for \texttt{two\_sided\_closure}.
Its rationale identifies incomplete arguments in the head's traces, motivating separate derivations for constructing an attainable answer and establishing the constraints that every answer must satisfy.
These proposals preserve free-form reasoning while developing new ways to construct and check solutions.

Later proposals extend this sequence through successive heads.
At iteration~17, the rehearsal harness preserves the two-sided structure and adds a preliminary attempt at a retrieved problem, using feedback from its reference solution to guide the target derivation.
At iteration~18, its successor adds a concrete challenge when the construction and necessity arguments agree, targeting errors shared by both paths.
At iteration~19, \texttt{gap\_directed\_construction} extends that head's disagreement path by attempting the missing construction before resolving the competing answers.
Its rationale explicitly preserves the retrieval, rehearsal, two-sided derivation, and agreement checks inherited from the head.
These recorded dependencies show how the transition at iteration~13 is followed by a sequence of harnesses that retain and extend mechanisms developed by subsequent heads.

\section{Proposal-Guidance Updates}
\label{app:branch-guidance}

\subsection{Full Update Prompts}
\label{app:skill-update-prompt}

We use the same proposer $P$, Claude Opus 4.6, to optimize branch-specific proposal guidance $S_b^t$, stored in \texttt{SKILL.md}.
The following prompts ask $P$ to update this file using local search history.

\paragraph{Mathematical reasoning}

\noindent\textbf{Update instructions.}
\begin{lstlisting}[basicstyle=\ttfamily\scriptsize,breaklines=true,columns=fullflexible,keepspaces=true]
---
name: branch-skill-gradient
description: Update one Math branch-local proposer skill from a completed-iteration window.
---

# Periodic Skill Gradient

Update the selected Math branch's local proposer skill using exactly the
iteration summaries supplied in the prompt.

## Constraints

- Read every supplied summary and do not seek older evolution history.
- Use validation accuracy as evidence. Compare within the same validation stage;
  cross-stage scores are reference only.
- Edit only the supplied branch-local `meta_skill/SKILL.md` and
  `meta_skill/gradient.json`.
- Do not edit agents, validation data, benchmark output, or source files.
- Preserve the proposer contract of exactly one candidate per iteration.
- Do not add problem text, answers, case IDs, or validation-specific rules.
- Treat each score as evidence about one implementation, not proof about a whole
  retrieval or reasoning approach.
- Use strong claims only when distinct implementations agree across at least two
  supplied iterations. Missing evaluations are missing data.
- Keep the skill open to evidence-informed exploration, not just exploitation.

## Method

1. Compare all candidate outcomes in the supplied iterations.
2. Separate candidate-specific facts from tentative broader hypotheses.
3. Give the latest validation stage the most weight.
4. State one objective for the next interval without collapsing exploration.
5. Write `gradient.json` per `gradient.schema.json` (validated fields:
   `completed_iteration`, `branch_id`, `objective`, non-empty `evidence`,
   non-empty `skill_changes`), then apply those calibrated changes to `SKILL.md`.
\end{lstlisting}

\paragraph{Terminal-Bench 2.0}

\noindent\textbf{Update instructions.}
\begin{lstlisting}[basicstyle=\ttfamily\scriptsize,breaklines=true,columns=fullflexible,keepspaces=true]
---
name: branch-skill-gradient
description: Update one branch-local TB2 proposer skill from a completed-iteration window.
---

# Periodic Skill Gradient

Update the selected TB2 branch's local proposer skill using exactly the
iteration summaries supplied in the prompt.

## Constraints

- Read every supplied summary and do not seek older evolution history.
- Use validation pass rate as evidence. Compare within the same validation stage;
  cross-stage scores are reference only.
- Edit only the supplied branch-local `meta_skill/SKILL.md` and
  `meta_skill/gradient.json`.
- Do not edit agents, validation task lists, benchmark output, or source files.
- Preserve the proposer contract of exactly one candidate per iteration.
- Do not encode task instructions, expected outputs, or task-specific solution steps.
- Treat each agent result as evidence about one implementation, not proof about a
  whole scaffold approach.
- Use strong claims only when distinct implementations agree across at least two
  supplied iterations. Missing evaluations are missing data.
- Keep the skill open to evidence-informed exploration, not just exploitation.

## Method

1. Compare all candidate outcomes in the supplied iterations.
2. Separate candidate-specific facts from tentative broader hypotheses.
3. Give the latest validation stage the most weight.
4. State one objective for the next interval without collapsing exploration.
5. Write `gradient.json` per `gradient.schema.json` (validated fields:
   `completed_iteration`, `branch_id`, `objective`, non-empty `evidence`,
   non-empty `skill_changes`), then apply those calibrated changes to `SKILL.md`.
\end{lstlisting}

\subsection{Concrete Gradient-Update Examples}
\label{app:gradient-examples}

\begin{figure}[!htbp]
\centering
\includegraphics[width=\textwidth]{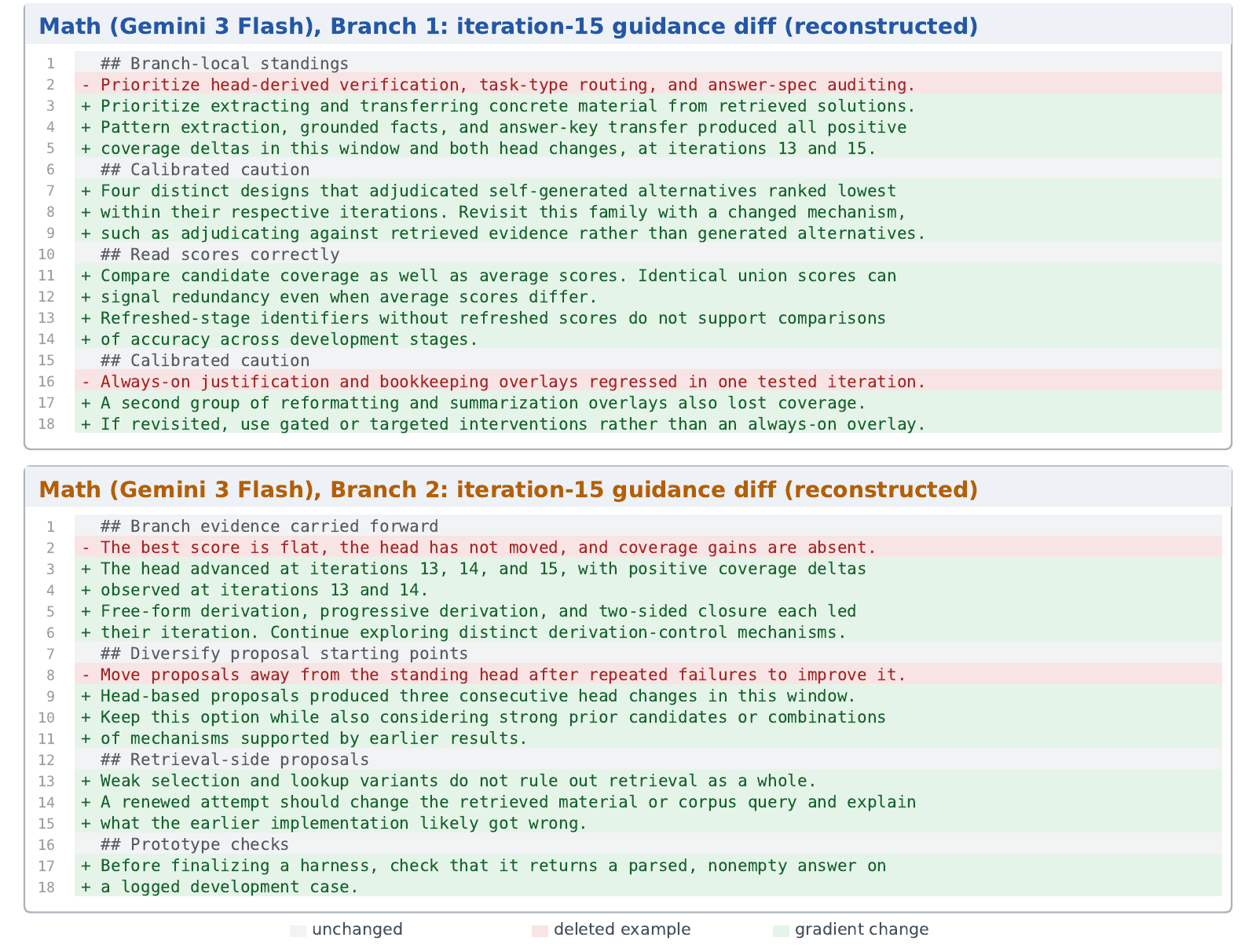}
\caption{Branch-specific \texttt{SKILL.md} changes on Math--Gemini from iteration-15 gradients.}
\label{fig:meta-skill-diff-gemini}
\end{figure}

\begin{figure}[!htbp]
\centering
\includegraphics[width=\textwidth]{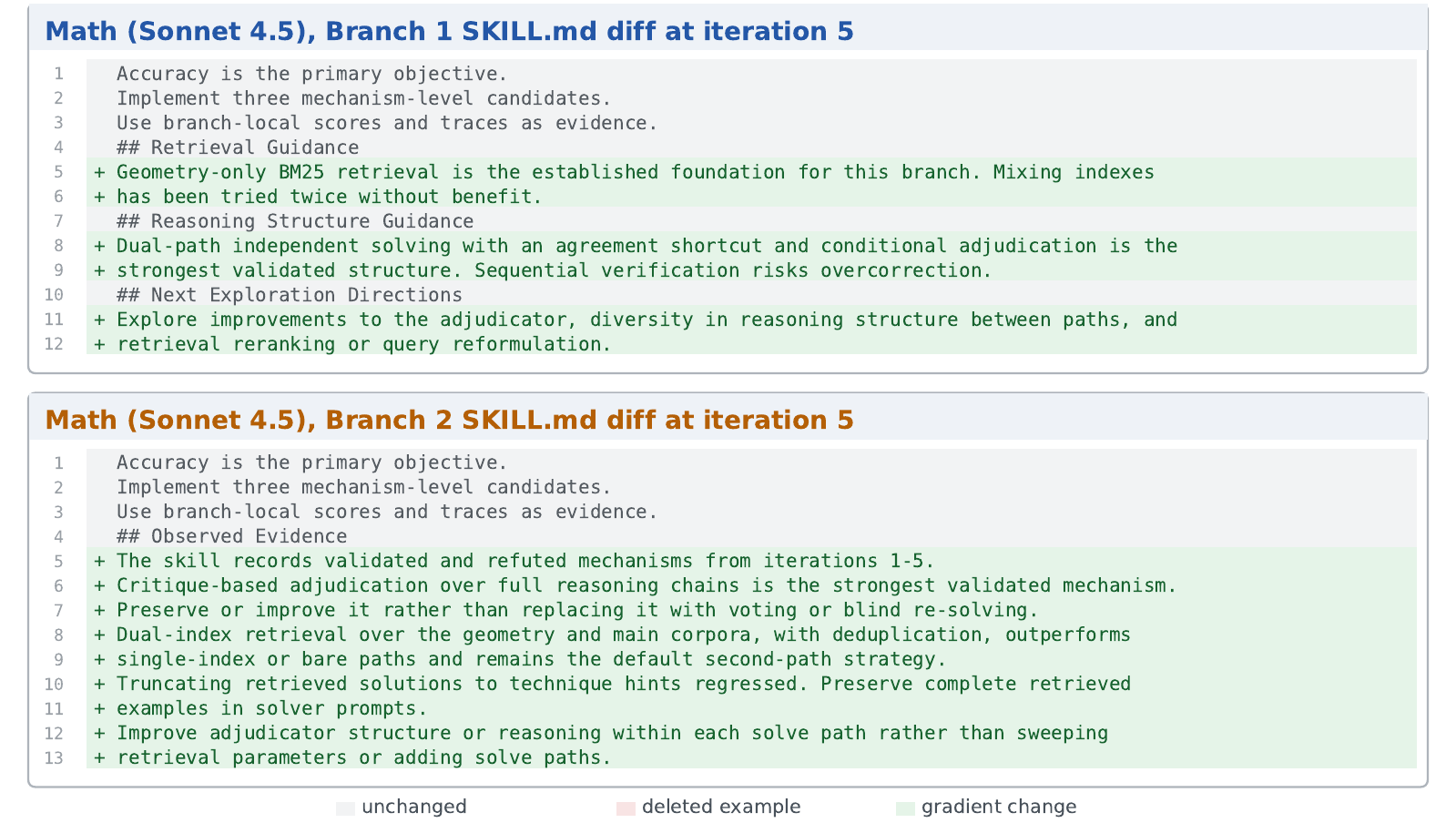}
\caption{Branch-specific \texttt{SKILL.md} changes on Math--Sonnet from iteration-5 gradients.}
\label{fig:meta-skill-diff-sonnet}
\end{figure}

\begin{figure}[!htbp]
\centering
\includegraphics[width=\textwidth]{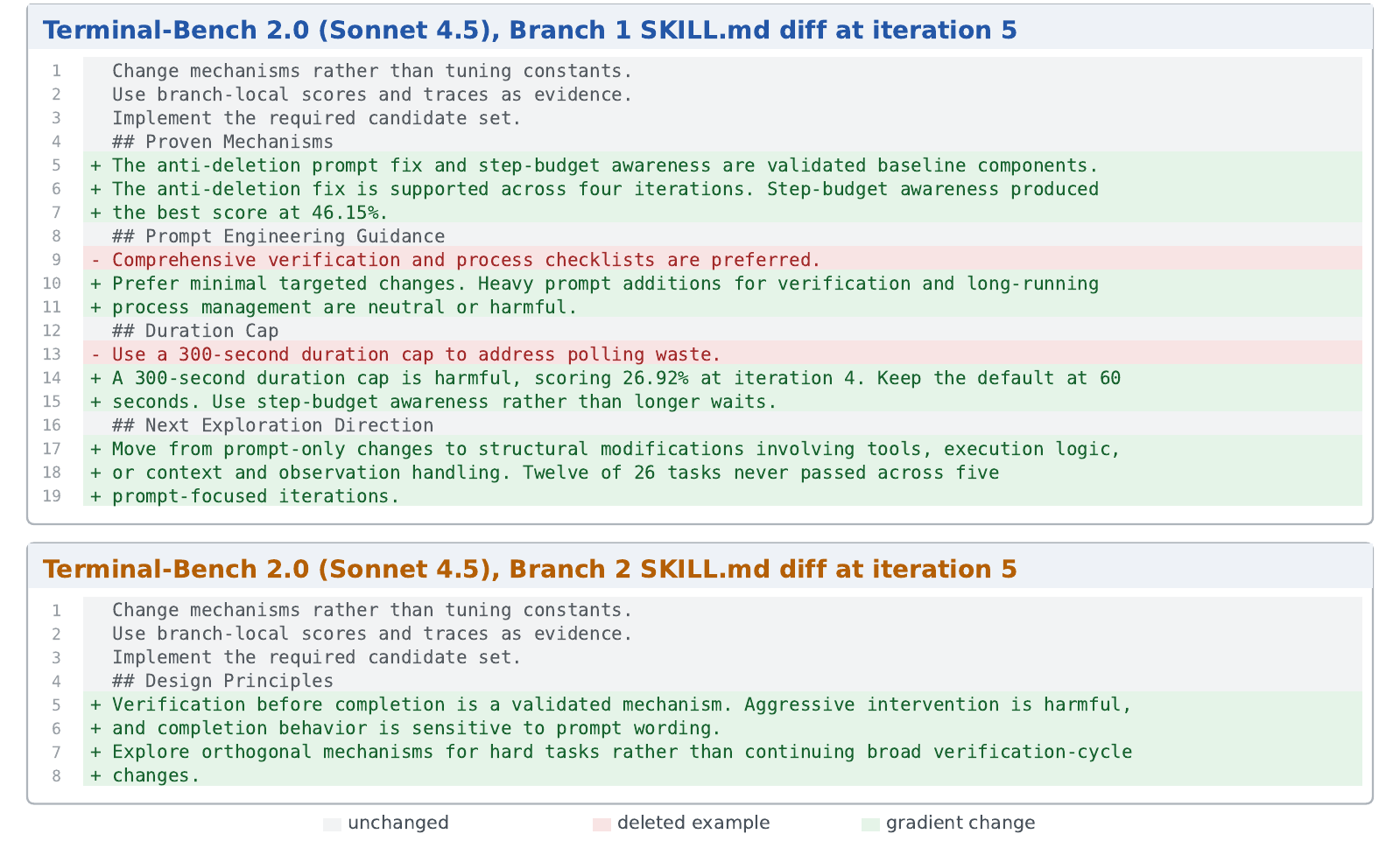}
\caption{Branch-specific \texttt{SKILL.md} changes on Terminal-Bench~2.0 from iteration-5 gradients.}
\label{fig:meta-skill-diff-tb2}
\end{figure}

\begin{figure}[!htbp]
\centering
\includegraphics[width=\textwidth]{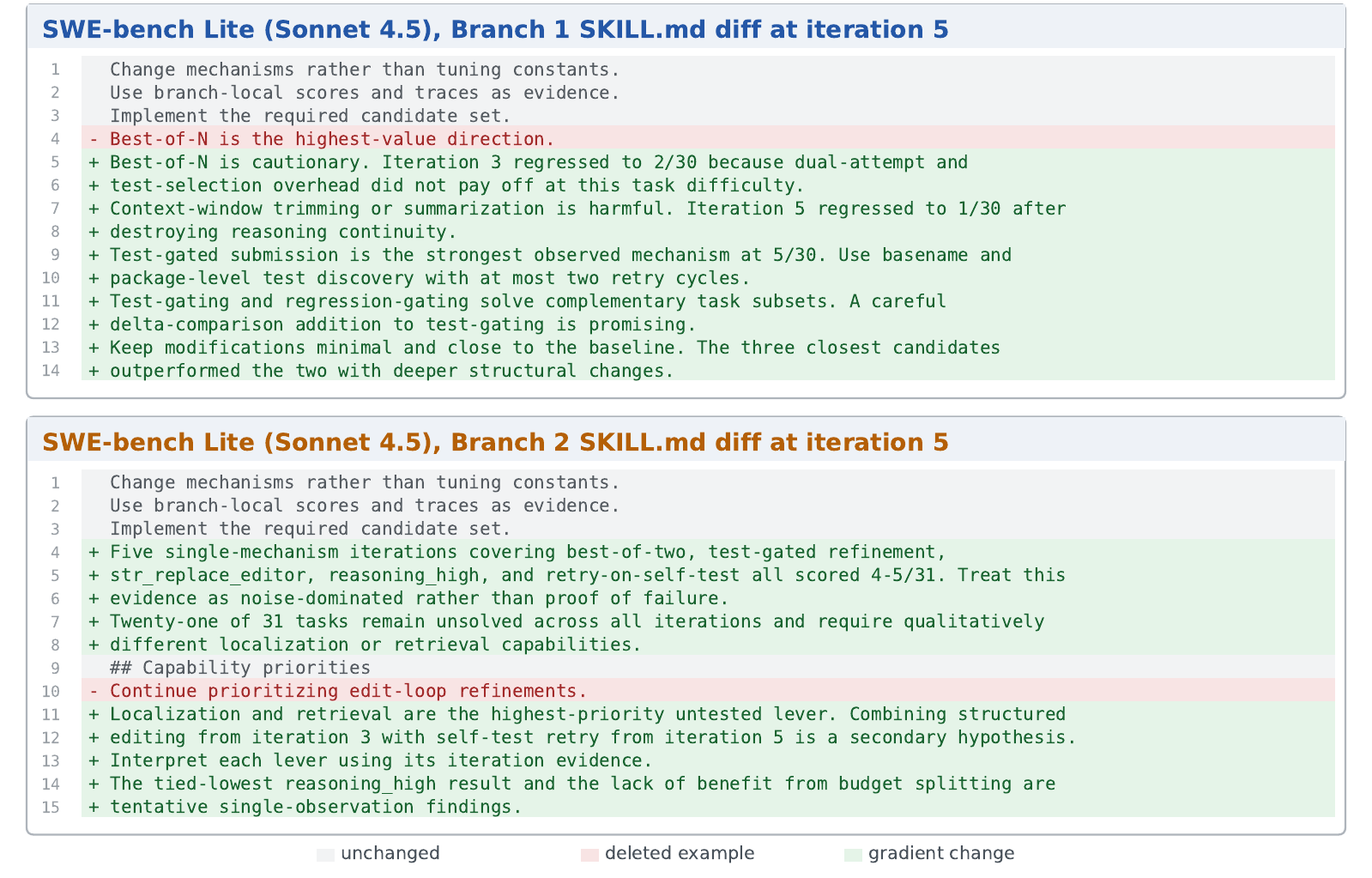}
\caption{Branch-specific \texttt{SKILL.md} changes on SWE-bench Lite from iteration-5 gradients.}
\label{fig:meta-skill-diff-swe}
\end{figure}

Using the prompts in Appendix~\ref{app:skill-update-prompt}, $P$ revises $S_b^t$ from local search evidence.
We present these updates as diffs of \texttt{SKILL.md}, with gray for context, red for deletions, and green for additions.
Figures~\ref{fig:meta-skill-diff-gemini}--\ref{fig:meta-skill-diff-swe} illustrate branch-specific guidance updates across Math--Gemini, Math--Sonnet, Terminal-Bench 2.0, and SWE-bench Lite.

\section{Router Analysis}
\label{app:router-analysis}

\subsection{Router Instructions}
\label{app:router-instructions}

The full router input combines the routing instruction optimized by GEPA, expert source code, and head-exclusive development cases with both experts' outputs.
The template below repeats the example block for each supplied development case, using solutions for math and final actions for Terminal-Bench 2.0 and SWE-bench Lite.

\begin{lstlisting}[basicstyle=\ttfamily\scriptsize\color{black},breaklines=true,columns=fullflexible,keepspaces=true]
{routing_instruction}

OUTPUT CONTRACT (follow exactly): return one JSON object with exactly two fields:
{"choice": "expert_1", "reason": "short evidence-based reason"}
or
{"choice": "expert_2", "reason": "short evidence-based reason"}
The choice must be the literal string "expert_1" or "expert_2".
Return no text outside the JSON object.

EXPERT 1 ({expert_1_name}) IMPLEMENTATION:
{expert_1_source_code}

EXPERT 2 ({expert_2_name}) IMPLEMENTATION:
{expert_2_source_code}

DECISION EXAMPLES (owned development cases):
EXAMPLE {i}:
PROBLEM:
{development_case}
EXPERT 1 {SOLUTION or FINAL ACTIONS}:
{expert_1_output_on_development_case}
EXPERT 2 {SOLUTION or FINAL ACTIONS}:
{expert_2_output_on_development_case}
WINNER: {EXPERT 1 or EXPERT 2} solved this; the other did not.

PROBLEM:
{new_problem}

Return JSON only: {"choice":"expert_1"|"expert_2","reason":"..."}
\end{lstlisting}

\subsection{Router Context and Prompt Ablations}
\label{app:router-context}
Table~\ref{tab:router-context} compares three input configurations, each with and without GEPA-based prompt optimization, across all four settings.
All configurations receive expert source code and select an expert before execution on the new problem.
Among configurations without GEPA, adding head-exclusive development cases changes accuracy from 60.5\% to 57.0\% on Math--Gemini, from 30.5\% to 30.0\% on Math--Sonnet, and from 46.6\% to 48.3\% on Terminal-Bench 2.0.
Adding expert responses then raises Math--Gemini accuracy to 59.5\% but lowers accuracy on Math--Sonnet and Terminal-Bench 2.0 to 29.0\% and 46.6\%, respectively.
All three configurations achieve 66.4\% on SWE-bench Lite.
Additional context therefore helps in some comparisons, but its benefits vary across settings.

GEPA~\citep{agrawal2025gepa} optimizes the routing instruction using head-exclusive development cases as training data, labeled by the successful final head.
The first two rows of Table~\ref{tab:router-context} compare its effect when both development examples and expert outputs are available.
GEPA increases accuracy from 59.5\% to 62.0\% on Math--Gemini, from 29.0\% to 30.5\% on Math--Sonnet, and from 46.6\% to 50.0\% on Terminal-Bench 2.0, while reducing SWE-bench Lite accuracy from 66.4\% to 66.0\%.
Averaged across benchmarks, GEPA yields small gains in all three input configurations, with the largest benefit from full context.
Table~\ref{tab:main-results} reports this full-context configuration with GEPA.

\begin{table}[t]
\centering
\caption{Router configurations and held-out performance (\%).
All variants receive expert source code.
Checkmarks indicate included context and whether GEPA optimization is applied.}
\label{tab:router-context}
\small
\setlength{\tabcolsep}{4pt}
\renewcommand{\arraystretch}{1.12}
\begin{tabular*}{\textwidth}{@{\extracolsep{\fill}}ccccccc@{}}
\toprule
\multicolumn{3}{c}{\textbf{Router components}} & \multicolumn{2}{c}{\textbf{Math}} & \textbf{Terminal-Bench 2.0} & \textbf{SWE-bench Lite} \\
\cmidrule(lr){1-3}\cmidrule(lr){4-5}\cmidrule(lr){6-6}\cmidrule(l){7-7}
\shortstack{Exclusive\\Cases} & \shortstack{Expert\\Outputs} & GEPA & \shortstack{Gemini 3\\Flash} & \shortstack{Claude Sonnet\\4.5} & \shortstack{Claude Sonnet\\4.5} & \shortstack{Claude Sonnet\\4.5} \\
\midrule
$\checkmark$ & $\checkmark$ & $\checkmark$ & \textbf{62.0} & 30.5 & \textbf{50.0} & 66.0 \\
$\checkmark$ & $\checkmark$ & --- & 59.5 & 29.0 & 46.6 & \textbf{66.4} \\
$\checkmark$ & --- & $\checkmark$ & 57.0 & \textbf{31.0} & 48.3 & \textbf{66.4} \\
$\checkmark$ & --- & --- & 57.0 & 30.0 & 48.3 & \textbf{66.4} \\
--- & --- & $\checkmark$ & 59.5 & 30.5 & 48.3 & 66.0 \\
--- & --- & --- & 60.5 & 30.5 & 46.6 & \textbf{66.4} \\
\bottomrule
\end{tabular*}
\end{table}

\section{Discovered Harness Mechanisms}
\label{app:harness-analysis}

This appendix describes the harnesses underlying our test-best results in Table~\ref{tab:test-best-results}, selected from the ten-candidate pools defined in Section~\ref{sec:test-reference}.
Each description follows the execution sequence and identifies the main change from the preceding design.

\subsection{Math--Gemini: Construction Before Adjudication}
\noindent\textbf{Overview.}
The \texttt{gap\_directed\_construction} harness attains the 58.0\% test-best score in the development-ranked pool.
It separates finding an attainable answer from establishing why no other answer is possible, then uses the relationship between these two arguments to determine the next computation.
The execution proceeds in three stages:
\begin{enumerate}
\item \textbf{Retrieve and rehearse.}
The harness generates a plan and uses it with the problem statement to retrieve worked examples through BM25.
When a suitable reference is available, it attempts that example and compares its attempt with the reference solution to produce a calibration note for the target problem.
\item \textbf{Construct and constrain.}
Two solver calls receive the plan, examples, and calibration note.
One constructs and checks an attainable answer, while the other derives necessary conditions on the answer.
\item \textbf{Resolve agreement or disagreement.}
If the normalized answers agree, a further call challenges the shared answer.
If they disagree, an additional call attempts an explicit construction or concrete check that addresses the gap before a final call resolves the problem.
\end{enumerate}
The construction step gives the adjudicator concrete evidence for resolving disagreements.

\subsection{Math--Sonnet: Cleaned Retrieval with Answer Grounding}
\noindent\textbf{Overview.}
The \texttt{clean\_retrieval\_answer\_grounded} harness achieves 32.5\% test accuracy using two retrieval-conditioned solution paths and conditional critique.
Its main change is how retrieved solutions are prepared before they enter the solver context.
\begin{enumerate}
\item \textbf{Retrieve complementary references.}
One path retrieves geometry examples, while the other retrieves distinct examples from the general corpus.
\item \textbf{Clean and ground the examples.}
The harness removes reasoning tags, trims tagged reasoning at a sentence boundary when one is available, caps each solution at 4,000 characters, and appends the reference answer stored in the corpus.
\item \textbf{Solve and critique.}
Both paths follow structured planning, solving, and verification instructions.
Matching normalized answers are returned directly.
On disagreement, a third call critiques both reasoning traces and produces the final answer.
\end{enumerate}
It preserves the two- or three-call solver structure while changing the retrieved context.

\subsection{Terminal-Bench 2.0: Verification Once Before Completion}
\noindent\textbf{Overview.}
The \texttt{verification\_once\_gate} harness achieves 50.0\% task completion.
It changes the completion checks of a verification-enhanced Terminus-KIRA agent.
\begin{enumerate}
\item \textbf{Execute the task.}
The agent interacts with the terminal through the inherited tool-use loop.
\item \textbf{Request verification once.}
When the agent first signals completion, the harness issues a verification prompt and records that the prompt has been shown.
\item \textbf{Allow subsequent completion.}
A later completion signal bypasses the same gate, allowing execution to finish instead of repeatedly requesting verification.
\end{enumerate}
The flag remains set during subsequent tool use, preventing the verification gate from reopening after the agent resumes work.

\subsection{SWE-bench Lite: Issue-Guided Localization Before Editing}
\noindent\textbf{Overview.}
The \texttt{test\_aware\_edit} harness achieves 66.4\% issue resolution.
It extends a structured-editing agent with a localization stage before the first model-driven action.
\begin{enumerate}
\item \textbf{Extract issue identifiers.}
The harness collects class names, function names, Python file paths, and exception names from the issue description, retaining up to eight identifiers.
\item \textbf{Assemble repository context.}
It searches Python files for these identifiers, separates source and test files, and ranks them by identifier matches.
It reads the top two source files and the top test file, caps their contents at 4,000 characters per source file and 3,000 for the test file, and inserts the context into the initial user message.
If no useful matches are found, the localization stage supplies no extra context.
\item \textbf{Run the editing loop.}
The agent continues with its existing tools, including a string-replacement operation that requires a unique match and returns the resulting diff.
\end{enumerate}
The retrieved repository tests provide examples of expected behavior before the agent begins editing.

\end{document}